\documentclass[lettersize,journal]{IEEEtran}
\usepackage{amsmath,amsfonts}
\usepackage{algorithmic}
\usepackage{algorithm}
\usepackage{array}
\usepackage[caption=false]{subfig}
\usepackage{textcomp}
\usepackage{stfloats}
\usepackage{url}
\usepackage{verbatim}
\usepackage{graphicx}
\usepackage{cite}
\usepackage{amsmath}
\usepackage{amssymb}
\usepackage{multirow}
\usepackage{colortbl}
\usepackage{booktabs}
\usepackage{makecell}
\usepackage{bbding}
\usepackage{setspace}
\usepackage{threeparttable}
\usepackage{balance}
\usepackage{tikz,xcolor}

\usepackage[colorlinks,
linkcolor=blue,
anchorcolor=blue,
citecolor=blue]{hyperref}

\hypersetup{hidelinks,
	colorlinks=true,
	allcolors=blue,
	pdfstartview=Fit,
	breaklinks=true}

\definecolor{lime}{HTML}{A6CE39}
\DeclareRobustCommand{\orcidicon}{
\begin{tikzpicture}
\draw[lime, fill=lime] (0,0)
circle[radius=0.16]
node[white]{{\fontfamily{qag}\selectfont \tiny \.{I}D}};
\end{tikzpicture}
\hspace{-2mm}
}
\foreach \x in {A, ..., Z}{%
\expandafter\xdef\csname orcid\x\endcsname{\noexpand\href{https://orcid.org/\csname orcidauthor\x\endcsname}{\noexpand\orcidicon}}
}

\begin{document}

%\title{A Sample Article Using IEEEtran.cls\\ for IEEE Journals and Transactions}
\title{Tracing Intricate Cues in Dialogue: Joint Graph Structure and Sentiment Dynamics for Multimodal Emotion Recognition}
%~\IEEEmembership{~IEEE,}
\author{Jiang Li\hspace{-1.5mm}\orcidA{}, Xiaoping Wang\hspace{-1.5mm}\orcidB{},~\IEEEmembership{Senior Member,~IEEE}, and Zhigang Zeng\hspace{-1.5mm}\orcidC{},~\IEEEmembership{Fellow,~IEEE}
        % <-this % stops a space
        
\thanks{Manuscript received 30 July 2024; revised 20 January 2025; revised 30 March 2025; accepted 16 June 2025. This work was supported in part by the Interdisciplinary Research Program of HUST under Grant 2024JCYJ006, in part by the National Natural Science Foundation of China under Grant 62236005 and 623B2040, in part by the Foundation for Outstanding Research Groups of Hubei Province of China under Grant 2025AFA012, in part by the 111 Project on Computational Intelligence and Intelligent Control under Grant B18024, and in part by the Fundamental Research Funds for the Central Universities of HUST under Grant 2024JYCXJJ067 and 2024JYCXJJ068. \textit{(Corresponding authors: Jiang Li and Xiaoping Wang.)}}
% \thanks{Manuscript received 30 July 2024; revised 20 January 2025; revised 30 March 2025; accepted 30 March 2025. This work was supported in part by the Interdisciplinary Research Program of HUST under Grant 2024JCYJ006, the National Natural Science Foundation of China under Grant 62236005 and 61936004, and the Fundamental Research Funds for the Central Universities of HUST under Grant 2024JYCXJJ067 and 2024JYCXJJ068. \textit{(Corresponding authors: Jiang Li and Xiaoping Wang.)}}
\thanks{The authors are with the School of Artificial Intelligence and Automation, Huazhong University of Science and Technology, Wuhan 430074, China, also with the Institute of Artificial Intelligence, Huazhong University of Science and Technology, Wuhan 430074, China, also with the Hubei Key Laboratory of Brain-Inspired Intelligent Systems, Huazhong University of Science and Technology, Wuhan 430074, China, and also with the Key Laboratory of Image Processing and Intelligent Control (Huazhong University of Science and Technology), Ministry of Education, Wuhan 430074, China (e-mail: lijfrank@hust.edu.cn; wangxiaoping@hust.edu.cn; zgzeng@hust.edu.cn).}
\thanks{Digital Object Identifier 10.1109/TPAMI.2025.3581236}
}

\markboth{IEEE TRANSACTIONS ON PATTERN ANALYSIS AND MACHINE INTELLIGENCE}
{LI \MakeLowercase{\textit{et al.}}: Tracing Intricate Cues in Dialogue: Joint Graph Structure and Sentiment Dynamics}

\IEEEpubid{0162--8828 ~\copyright~2025 IEEE. All rights reserved, including rights for text and data mining, and training of artificial intelligence and similar technologies. Personal use is permitted, but republication/redistribution requires IEEE permission. See https://www.ieee.org/publications/rights/index.html for more information.}

\maketitle

\begin{abstract}
Multimodal emotion recognition in conversation (MERC) has garnered substantial research attention recently. Existing MERC methods face several challenges: (1) they fail to fully harness direct inter-modal cues, possibly leading to less-than-thorough cross-modal modeling; (2) they concurrently extract information from the same and different modalities at each network layer, potentially triggering conflicts from the fusion of multi-source data; (3) they lack the agility required to detect dynamic sentimental changes, perhaps resulting in inaccurate classification of utterances with abrupt sentiment shifts. To address these issues, a novel approach named GraphSmile is proposed for tracking intricate emotional cues in multimodal dialogues. GraphSmile comprises two key components, i.e., GSF and SDP modules. GSF ingeniously leverages graph structures to alternately assimilate inter-modal and intra-modal emotional dependencies layer by layer, adequately capturing cross-modal cues while effectively circumventing fusion conflicts. SDP is an auxiliary task to explicitly delineate the sentiment dynamics between utterances, promoting the model's ability to distinguish sentimental discrepancies. GraphSmile is effortlessly applied to multimodal sentiment analysis in conversation (MSAC), thus enabling simultaneous execution of MERC and MSAC tasks. Empirical results on multiple benchmarks demonstrate that GraphSmile can handle complex emotional and sentimental patterns, significantly outperforming baseline models.
\end{abstract}


\section{Introduction}
\IEEEPARstart{E}{motion} recognition in conversation (ERC) has become a hot topic in academia due to its broad application prospects in various fields such as government affairs, education, and healthcare. A multitude of ERC models~\cite{jiao2019higru,ghosal2020cosmic,Ong_Su_Chen_2022,hu-etal-2023-supervised} focus on the analysis of textual modality, often encountering the issue of insufficient information. In certain special cases, such as sarcasm, text-based methods may not fully comprehend the semantics of an utterance from solely a single modality. The extraordinary ability of the human brain to integrate multisensory information provides inspiration for addressing the limitations of information deficiency in unimodal models. Multimodal emotion recognition in conversation (MERC) emerges as an effective solution to this challenge, achieving more accurate emotional state recognition by integrating information from various modalities such as text, audio, and video. MERC is a sophisticated task that goes beyond the scope of traditional unimodal~\cite{shen-etal-2021-directed,zhao2022cauain} and non-conversational~\cite{hazarika2020misa,mai2022hybrid} affective endeavors (in this paper, the term ``affective" encompasses emotional and sentimental aspects). It necessitates not only the sufficient extraction of cross-modal association cues that capture synergies between different sensory inputs (e.g., facial expressions, vocal tones, and textual content), but also an in-depth investigation of intra-modal contextual cues to perceive effects of contexts on the emotional state of an utterance. Compared to unimodal and non-conversational counterparts, MERC faces more formidable challenges, with the key one being the thorough capture of intra-modal contextual cues and inter-modal associative cues.

\IEEEpubidadjcol
Researchers have developed numerous MERC approaches, including new architectures and learning mechanisms, to capture and synthesize the wealth of information present in multimodal dialogues. Some efforts~\cite{majumder2019dialoguernn,li2022emocaps,hu2022unimse,10.1145/3581783.3612053} straightforwardly concatenated or summed features across disparate modalities, which have manifested certain performance merits over unimodal approaches when dealing with specific tasks. Regrettably, certain modalities may convey information that is ambiguous or even contradictory to labels. These concatenation fusion-based works fail to adequately account for the intrinsic relationships and interactions between different modalities, thus making it difficult to effectively address the issues of information selection and integration. There are an array of methods~\cite{mao2021dialoguetrm,9316758,zhang-li-2023-cross,zheng-etal-2023-facial} predicated on cross-attention fusion. 
These methods achieve the amalgamation of diverse modality information by calculating the weights between multimodal features and applying these weights to the current modality features. Nonetheless, they retain certain constraints at the feature level and have not fully actualized direct interaction and fusion between different modalities.

Recent studies have unveiled the exceptional potential of graph neural networks (GNNs) in handling the fusion and comprehension of multimodal data. GNNs, with the unique topological structure and dynamic information propagation mechanism, inherently excel at capturing cross-modal associative cues and intra-modal contextual cues. 
Despite significant advancements~\cite{hu2021mmgcn,10078161,10219015,10203083} in the graph-based MERC domain, several unresolved issues persist. Firstly, existing models, when constructing cross-modal graphs, only connect heterogeneous nodes (i.e., nodes that belong to different modalities) from the same utterance and fail to incorporate associations with heterogeneous nodes from different utterances. This construction scheme overlooks the direct emotional cues provided by other utterances across modalities, leading to incomplete capture of multimodal information. Secondly, these models typically process information from the same and different modalities at each layer of the network, i.e., concurrently extract intra-modal and inter-modal emotional cues. Owing to the heterogeneity among modalities~\cite{10.1145/3656580,10.1145/3649447,10078161}, this concurrent aggregation of different types of information may precipitate conflicts during fusion, resulting in insufficient cross-modal modeling. Furthermore, current models are excessively preoccupied with contextual modeling, neglecting the dynamic mutation of sentiments throughout the conversation. The absence of agility in recognizing sentimental changes may cause these models' inability to classify accurately in scenarios with abrupt sentiment shifts. Consequently, there is an urgent need to develop models that can integrate multimodal heterogeneous information more comprehensively and rationally, and that are capable of keenly capturing the sentiment dynamics.

To cope with the above dilemmas, we introduce a joint graph structure and sentiment dynamics for multimodal emotion recognition in conversation (GraphSmile). The core components of GraphSmile encompass the graph structure fusion (GSF) and sentiment dynamics perception (SDP) modules. During the construction of cross-modal edges, GraphSmile not only correlates the current node with heterogeneous nodes belonging to the same utterance but also directly connects it with nodes emanating from disparate utterances. This strategy for edge construction effectively ameliorates the insufficient in extracting cross-modal information. At the graph convolution stage, the GSF module adeptly utilizes the graph structure to alternately collect inter-modal associative cues and intra-modal contextual cues on layer-by-layer basis, thereby precluding potential conflicts stemming from the simultaneous aggregation of diverse data sources at each layer. To mitigate redundant connections in the multimodal dialogue graph, we refrain from constructing edges between nodes within the same modality; nonetheless, the GSF module, through its unique graph convolution mechanism, is capable of directly and comprehensively capturing intra-modal contextual cues. In the phase of sentiment dynamics detection, we integrate the SDP module as an auxiliary task to explicitly model the shift in sentimental state between utterances. This module not only deepens the model's comprehension of sentiment dynamics but also constrains the network parameters of the main task with a strategy akin to contrastive learning, i.e., drawing utterances with the same sentiment closer and those with different sentiments further apart, thereby enhancing the model's ability to discern sentimental disparities.

In this work, we can effortlessly extend GraphSmile into a multitask model. Namely, GraphSmile is applied to the multimodal sentiment analysis in conversation (MSAC) task and achieves multitask joint optimization with the MERC task. Specifically, we initially incorporate a sentiment classifier, followed by the joint formulation of emotion classification loss, sentiment classification loss, and shift classification loss to construct a multitask learning objective, thereby synergistically optimizing network parameters. In this manner, our GraphSmile is capable of efficiently executing both MSAC and MERC tasks. In the domain of affective computing, emotions (e.g., happiness, surprise, neutrality, sadness, and anger) are typically regarded as more nuanced and fleeting psychological states, characterized by their situational and temporary nature, while sentiments (e.g., negative, neutral, positive) are understood as coarser and more enduring psychological inclinations, marked by their profundity and stability. While emotion and sentiment exhibit varying levels of granularity, duration, and characteristic, there exists an interrelation between them; happiness, for example, is typified as a positive sentiment, whereas sadness is reflective of a negative sentimental disposition.
Thus, our co-optimization strategy not only facilitates knowledge transfer and complementary advantages between different granularity levels of affective tasks but also significantly enhances GraphSmile's comprehensive perceptive capability for emotional and sentimental cues.

The principal contributions of this work are as follows:
\begin{itemize}
  \item A multitask multimodal affective model that unites graph structure and sentiment dynamics, GraphSmile, is proposed for both MERC and MSAC tasks. GraphSmile exploits multiple losses to achieve multitask synergistic optimization while upgrading its ability to synthesize the perception of various cues.
  \item The GSF module we designed not only ensures the full exploitation of cross-modal affective cues but also obviates potential conflicts arising from the fusion of multi-source data. This module also confers two additional benefits, i.e., it diminishes redundant connections within the graph and alleviates the over-smoothing issue prevalent in graph-based models.
  \item We develop a plug-and-play SDP module that can explicitly capture sentimental changes in dialogues, fostering the model's understanding of the dynamics of sentimental evolution while constraining the network parameters of the primary task.
  \item Extensive comparative experiments and ablation studies have been conducted on multimodal benchmark datasets. The empirical results have validated the superiority of GraphSmile in both MERC and MSAC tasks.
\end{itemize}

The remainder of this paper is organized as follows. Section~\ref{sec:related_work} reviews the related efforts. Section~\ref{sec:methodology} provides a detailed exposition of the proposed GraphSmile. Sections~\ref{sec:experimental_setup} and~\ref{sec:experimental_result} constitute the experimental segment of our work. Section~\ref{sec:conclusion} summarizes this work.

\section{Related Work}\label{sec:related_work}
\subsection{Emotion Recognition in Conversation}
With the proliferation of applications for dialogue systems, ERC has increasingly become a focal point in the realm of affective computing. Depending on the source of input information, research in this domain can be categorized into two branches: unimodal and multimodal approaches. Since the inception of ERC tasks, text-based methods~\cite{Li_Yan_Qiu_2022,song-etal-2022-supervised,tu-etal-2023-context,10.24963/ijcai.2023/699,yu-etal-2024-emotion} have consistently held a dominant position among unimodal approaches. In recent years, this branch has seen profound advancements with numerous emerging technologies modeling contextual information in textual conversations from diverse perspectives. 
GNNs, due to their proficiency in capturing intra-modal dependency information, have notably been instrumental in conversational relationship modeling, thereby giving rise to numerous graph-based ERC models~\cite{ghosal2019dialoguegcn,wang2020relational,lee-choi-2021-graph,ren2022lrgcn,zhang-etal-2023-dualgats,10418539}.
Regrettably, unimodal ERC methods are limited by their inherent information deficiencies and susceptibility to external influences, which can impede their performance. 

The richness of human emotions manifests distinctively across individual modalities such as text, audio, and vision. These modalities possess unique characteristic and exhibit both shared traits and individual nuances. Consequently, multimodal ERC techniques have garnered significant attention due to their capacity to integrate diverse modal information. 
Some methods~\cite{majumder2019dialoguernn,li2022emocaps,hu2022unimse,10008078,10.1145/3581783.3612053} employ a kind of straightforward feature fusion strategy, which involves concatenating or summing feature vectors from different modalities, overlooking the intrinsic interactions between modalities. In tandem, a variety of models~\cite{9316758,chudasama2022m2fnet,ZOU2022109978,zheng-etal-2023-facial} based on cross-modality attention are proposed in succession.
These methods are unable to achieve direct absorption of information from one modality to another due to the constraints of their network structures, which are considered implicit cross-modal modeling.

In recent years, GNNs have garnered widespread attention due to their powerful capabilities in handling complex relationships, providing a solvable inspiration for explicitly conducting cross-modal interactions. 
Despite certain advancements in graph-based fusion methods~\cite{hu2022mmdfn,10078161,10219015,Tu_Xie_Liang_Wang_Xu_2024}, they still face a series of unique and common issues: due to the limitations of graph construction strategies, they may not fully capture and utilize cross-modal emotional cues; the heterogeneity between different modal data may lead to conflicts in the fusion process of multi-source information; some methods may overly rely on extracting intra-modal contextual cues, neglecting the dynamic changes in sentimental states during the conversation. Moreover, some of the aforementioned issues also exist in methods based on concatenation fusion and cross-attention fusion. In response to these issues, we propose a new fusion network architecture called GraphSmile for MERC.

\subsection{Multimodal Sentiment Analysis}
Multimodal sentiment analysis (MSA), due to its wide range of application prospects, has become a focal research topic in the field of affective computing. MSA is a multimodal task aimed at understanding and detecting human emotional tendencies by processing multimodal heterogeneous data, such as text, speech, and vision. The main challenge of the MSA task is how to effectively harness information from different modalities to complement each other, and a plethora of representative works~\cite{mai-etal-2019-divide,han-etal-2021-improving,10122560,10587106} have been proposed in succession. Existing mainstream works focus on enhancing the performance of MSA models by designing complex structures, interaction mechanisms, or fusion paradigms, which can be broadly categorized into tensor-based methods~\cite{mai-etal-2019-divide,9338320}, attention-based methods~\cite{tsai-etal-2019-multimodal,10122560}, and graph-based methods~\cite{yang-etal-2021-mtag,rrepo34855}. These research trajectories of MSA tasks are distinctly different from those of the ERC tasks, as they do not take into account the conversational context, rendering them incompatible with ERC tasks. Although Hu et al.~\cite{hu2022unimse} have made effort to unify ERC and MSA, their approach to handling labels in the datasets is overly complex and cannot fully guarantee the correctness of universal labels. Divergent from the extant MSA methods, our GraphSmile is conceived as a multitask multimodal conversational affective model. In GraphSmile, we only need to merge emotional labels to effortlessly unify the ERC and MSA tasks, thereby facilitating an efficacious excavation of both emotional and sentimental cues in a concurrent manner.

\section{Proposed GraphSmile}\label{sec:methodology}
Given a conversation $c = [u_1, u_2, \ldots, u_M]$ consisting of $M$ utterances, along with the emotional label $e_i$ (or sentimental label $s_i$) corresponding to each utterance $u_i$, the objective of the MERC (or MSAC) task is to train a model based on the dialogue and labels to detect the affective state of each utterance. In MERC and MSAC tasks, each utterance comprises three modalities: textual $t$, visual $v$, and acoustic $a$ modalities, which is denoted as $u_i = [u_i^t, u_i^v, u_i^a]$.

Effectively extracting complex affective cues from multimodal dialogues is crucial for the MERC task. Here, the affective cues mainly consist of inter-modal associative and intra-modal contextual cues. \textit{Inter-Modal Associative Cues:} Assuming the existence of two modalities (e.g., textual and visual), an utterance $u_i$ is represented differently across modalities as $u_i^t$ and $u_i^v$. The information carried by $u_i^t$ is considered as an inter-modal associative cue for $u_i^v$. The inter-modal associative cues for $u_i^v$ also include information carried by other utterances from modality $t$, such as $u_{i-2}^t$, $u_{i-1}^t$, and $u_{i+1}^t$. \textit{Intra-Modal Contextual Cues:} Taking a single modality (e.g., acoustic) as an example, an utterance $u_j$ is denoted as $u_j^a$ in a specific modality. The intra-modal contextual cues for $u_j^a$ encompass information carried by other utterances from the same modality $a$, such as $u_{j-1}^a$, $u_{j+1}^a,$ and $u_{j+2}^a$.

\begin{figure*}[htbp]
\centering
\includegraphics[width=\linewidth]{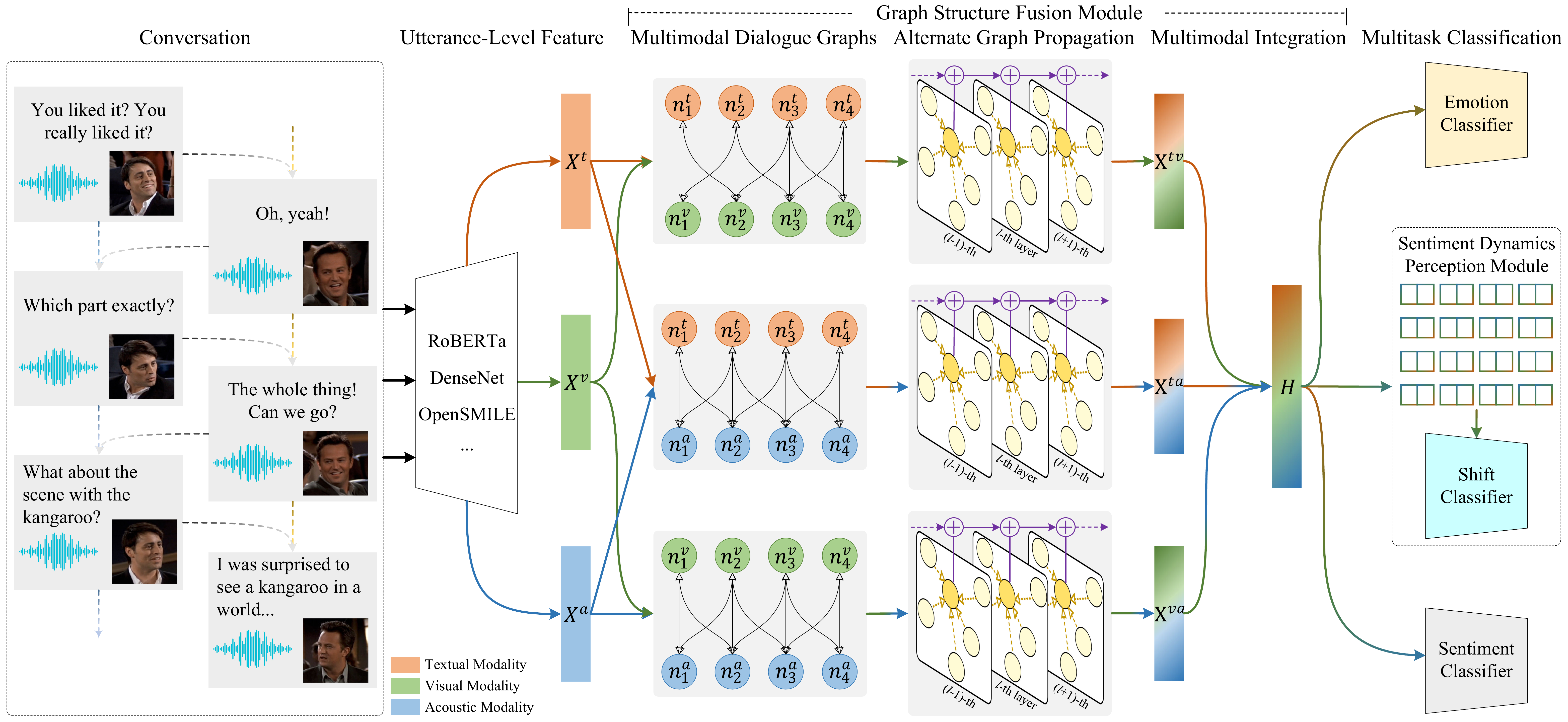}
\caption{Overall architecture of our GraphSmile. Initially, utterance-level features are extracted for each modality. Subsequently, three multimodal dialogue graphs are constructed. Then, alternating graph propagation is conducted to explore affective cues. Finally, multitask classification is performed.}
\label{fig:overall_architecture}
\end{figure*}

To address the predicaments encountered by existing models, we propose a model termed GraphSmile that conducts a comprehensive modeling of multimodal dialogues. The overall architecture of GraphSmile is depicted in Fig.~\ref{fig:overall_architecture}, which mainly includes the extraction of utterance-level features, the construction of multimodal dialogue graphs, the alternate graph propagation, the multimodal integration, and the multitask classification. Here, multitask classification involves emotion, shift, and sentiment classifications. In the following, we describe each component of GraphSmile in detail.

\subsection{Utterance-Level Feature}
The datasets utilized in this work include IEMOCAP~\cite{busso2008iemocap} (both 6-way and 4-way settings), MELD~\cite{poria2019meld}, and CMU-MOSEI~\cite{bagher-zadeh-etal-2018-multimodal}. Influenced by prior works~\cite{ghosal2020cosmic,hu2021mmgcn,joshi-etal-2022-cogmen}, the utterance-level feature extraction for the datasets used in GraphSmile is as follows. Text features for all datasets are processed using RoBERTa~\cite{liu2019roberta}. 
Visual features for the IEMOCAP (6-way) and MELD datasets are extracted using DenseNet~\cite{huang2017densely}, which is pre-trained on the FER+ corpus~\cite{barsoum2016training}. The IEMOCAP (4-way) dataset utilizes OpenFace 2.0~\cite{8373812} for visual feature extraction, while the visual features for the CMU-MOSEI dataset are derived from the work of Zadehet et al.~\cite{bagher-zadeh-etal-2018-multimodal}.
Audio features for the IEMOCAP (6-way) and MELD datasets are extracted using the OpenSmile toolkit with IS10 configuration~\cite{SCHULLER20111062}. The IEMOCAP (4-way) dataset uses OpenSmile~\cite{10.1145/1873951.1874246} for audio feature extraction, and the CMU-MOSEI dataset uses librosa~\cite{brian_mcfee-proc-scipy-2015} for this purpose.

\subsection{Graph Structure Fusion Module}
\subsubsection{Multimodal Dialogue Graphs}
To establish relationships between utterances while considering inter-modal interactions, we construct three multimodal dialogue graphs, each composed of two modalities. Taking the textual and visual modalities as an example, we build a textual-visual dialogue graph $\mathcal{G}^{tv} = (\mathcal{N}^t, \mathcal{N}^v, \mathcal{E}^{t \leftarrow v}, \mathcal{E}^{v \leftarrow t})$. Here, $\mathcal{N}^t$ and $\mathcal{N}^v$ represent the node sets from the textual and visual modalities, respectively. $\mathcal{E}^{t \leftarrow v}$ denotes the set of edges from source visual nodes to target textual nodes, while $\mathcal{E}^{v \leftarrow t}$ denotes the set of edges from source textual nodes to target visual nodes.

\textit{Nodes:} Each utterance $u_i$ is represented as nodes $n_i^t$ for the textual modality and $n_i^v$ for the visual modality, with corresponding features denoted as $x_i^t$ and $x_i^v$. If a dialogue consists of $m$ utterances, then $\mathcal{G}^{tv}$ contains $2m$ nodes. Consequently, $\mathcal{N}^t$ can be represented as $\{n_1^t, n_2^t, \ldots, n_M^t\}$, $\mathcal{N}^v$ can be represented as $\{n_1^v, n_2^v, \ldots, n_M^v\}$, with their corresponding features denoted as $[x_1^t, x_2^t, \ldots, x_M^t]$ and $[x_1^v, x_2^v, \ldots, x_M^v]$.

\textit{Edges:} We establish relationships only between nodes of different modalities. In the graph $\mathcal{G}^{tv}$, a textual node $n_i^t$ is connected to visual nodes; similarly, a visual node $n_i^v$ is connected to textual nodes. To mitigate redundant connections~\cite{ghosal-etal-2019-dialoguegcn,10078161}, we set up two sliding windows. Specifically, $n_i^t$ is associated with the past $P$ visual nodes (i.e., $n_{i-P}^v, n_{i-P+1}^v, \ldots, n_{i-1}^v$), and also with the future $F$ visual nodes (i.e., $n_{i+1}^v, n_{i+2}^v, \ldots, n_{i+F}^v$). The same connection way is applied to the visual node $n_i^v$. Additionally, different modal nodes (e.g., $n_i^t$ and $n_i^v$) from the same utterance are also connected.

\textit{Edge Weights:} In our research, the edges connecting two nodes are directed, with edges of differing orientations signifying distinct implications. For instance, in the graph $\mathcal{G}^{tv}$, for nodes $n_i^t$ and $n_{i-1}^v$, the edge $e(n_i^t \leftarrow n_{i-1}^v)$ denotes the informational contribution from $n_{i-1}^v$ to $n_i^t$, whereas the edge $e(n_{i-1}^v \leftarrow n_i^t)$ signifies the informational contribution from $n_i^t$ to $n_{i-1}^v$. To distinguish between different edges, diverging from existing methods~\cite{hu2021mmgcn,10078161}, we assign a set of trainable weights directly to them. These weights are initialized to a value of 1 and are optimized during the model training process.

\textit{Adjacency Matrix:} The adjacency matrix serves to store the relationships between all nodes in a graph, representing a pivotal element in expressing the graph structure and also constituting one of the input data for spectral-based GNNs. Based on the nodes, edges, and edge weights previously established, an adjacency matrix $A^{tv}$ can be generated to serve as the input for our GraphSmile.

Following the construction manner of the graph $\mathcal{G}^{tv}$, a textual-acoustic dialogue graph $\mathcal{G}^{ta}$ and a visual-acoustic dialogue graph $\mathcal{G}^{va}$ can also be obtained. The majority of previous approaches~\cite{hu2021mmgcn,10078161,10203083}, when creating cross-modal edges, only connect nodes from the same utterance, thereby neglecting the affective cues directly provided by other utterances. In contrast to these methods, in our construction of cross-modal edges, the current node is not only connected to nodes from the same utterance but also directly linked to nodes from different utterances. It is noteworthy that our multimodal dialogue graphs don't establish edges between utterances in the same modality. Nonetheless, GraphSmile can still effortlessly capture intra-modal contextual affective cues through the unique graph convolution mechanism. In Part~\ref{sec:example_explanation}, we explain how these intra-modal affective cues can be naturally extracted by graph convolution.

\subsubsection{Alternate Graph Propagation}
Previous efforts~\cite{hu2021mmgcn,10078161,10203083} have demonstrated that GNNs can effectively mine intra-modal contextual information and cross-modal associative information in multimodal dialogues. However, these models share a common issue: the simultaneous integration of inter-modal and intra-modal affective cues at each network layer may cause conflicts among multi-source information, making thorough cross-modal modeling difficult. This part will provide a detailed description of how the GSF module cleverly addresses this issue. 

As depicted in Fig.~\ref{fig:overall_architecture}, drawing inspiration from existing GNN works, we define a new graph convolutional operation to extract affective cues from multimodal dialogues. For each dialogue graph, multiple simplified graph convolutional layers~\cite{pmlr-v97-wu19e} are utilized to propagate inter-modal and intra-modal cues layer by layer in an alternating fashion, which can be mathematically expressed as:
\begin{equation}
\label{eq:sgclayer}
X^{(l)} = A X^{(l-1)} \Theta^{(l)}, \quad l = 1, 2, \ldots, L.
\end{equation}
Here, $A$ denotes the adjacency matrix of the dialogue graph, $X^{(l)}$ signifies the output at the $l$-th layer, and $X^{(0)}$ represents the initial input to the GSF module; $\Theta^{(l)}$ indicates the learnable parameter at the $l$-th layer. Assuming a given textual-visual dialogue graph $\mathcal{G}^{tv}$, then the adjacency matrix $A$ can be represented by $A^{tv}$, $X^{(l)}$ can be denoted by $X^{tv, (l)}$. The initial feature $X^{tv, (0)}$ is composed of $X^{t, (0)}$ and $X^{v, (0)}$, where $X^{t, (0)}$ and $X^{v, (0)}$ are denoted as $[x_1^t, x_2^t, \ldots, x_M^t]$ and $[x_1^v, x_2^v, \ldots, x_M^v]$, respectively.

Subsequently, the feature representations obtained from all layers are summed together, which is known as residual connections, and can be mathematically formulated as:
\begin{equation}
\label{eq:res_connect}
\mathrm{X} = X^{(0)} + \mathtt{FC}^{(1)}(X^{(1)}) + \cdots + \mathtt{FC}^{(L)}(X^{(L)}).
\end{equation}
In the graph $\mathcal{G}^{tv}$, $\mathrm{X}$ can be represented as $\mathrm{X}^{tv}$, which is composed of $\mathrm{X}^{t \leftarrow v}$ and $\mathrm{X}^{v \leftarrow t}$. Here, $\mathrm{X}^{t \leftarrow v}$ denotes the textual feature representation that has incorporated visual information, and $\mathrm{X}^{v \leftarrow t}$ signifies the visual feature representation that has carried textual information. $\mathtt{FC} (\cdot)$ represents a fully connected layer, which is composed of a linear transformation, an activation function (e.g., LeakyReLU), and a dropout operation, i.e.,
\begin{equation}
\label{eq:fully_connect}
\mathtt{FC}(X) = \mathtt{Dropout}(\mathtt{LeakyReLU}(X \Theta_f + b_f)).
\end{equation}
Here, $\Theta_f$ and $b_f$ are the learnable parameters. In our GSF module, the fully connected layer serves two purposes: (1) the trainable network parameters allow the GSF module to dynamically select information from each network layer; (2) the combination of the linear transformation and activation function can further enhance the expressive ability of the designed GSF module.

\subsubsection{Example Explanation}\label{sec:example_explanation}
\begin{figure}[htbp]
\centering
\subfloat[Adjacency Matrix]{\includegraphics[width=0.32\columnwidth]{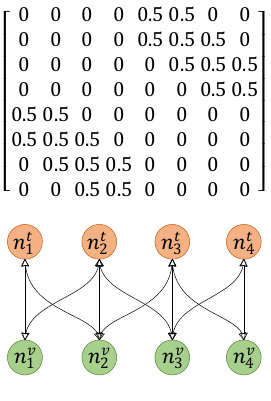}%
  \label{fig:example_power_1}}
\hfil
\subfloat[Matrix's Square]{\includegraphics[width=0.32\columnwidth]{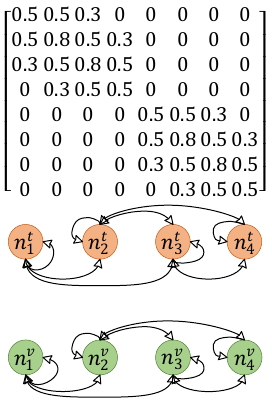}%
  \label{fig:example_power_2}}
\hfil
\subfloat[Matrix's Cube]{\includegraphics[width=0.32\columnwidth]{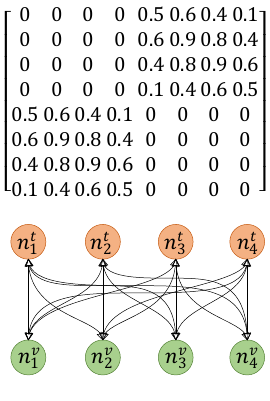}%
  \label{fig:example_power_3}}
\caption{Example explanation of alternating propagation in the GSF module. Here, (a), (b), and (c) represent the adjacency matrix $A$, the matrix's square $A^2$, and the matrix's cube $A^3$, along with their corresponding graph structures, respectively. In actual implementation, the edge weights in the dialogue graph $\mathcal{G}^{tv}$ are trainable parameters; for the sake of narrative simplicity, we assume all edge weights to be 0.5.}
\label{fig:example_power}
\end{figure}

We take the textual-visual dialogue graph $\mathcal{G}^{tv}$ as an example to elucidate why the GSF module can alternately propagate different types of affective cues layer by layer. In Fig.~\ref{fig:example_power}, we illustrate the process of multi-layer graph convolution with a concrete example involving four utterances.

According to Eq.~\ref{eq:sgclayer}, the first-layer graph convolution process can be mathematically formulated as: 
\begin{equation}
\label{eq:sgc1}
X^{(1)} = A X^{(0)} \Theta^{(1)}.
\end{equation}
Here, $A$ can be represented by $A^{tv}$, which is the adjacency matrix that expresses the relationships between all nodes in the graph $\mathcal{G}^{tv}$; $X^{(1)}$ denotes the output at the first layer, $X^{(0)}$ is the initial input to the GSF module, and $\Theta^{(1)}$ indicates the learnable parameter at the first layer. From the construction way of the multimodal dialogue graph (see Fig.~\ref{fig:example_power_1}), $A^{tv}$ does not include the relationships between nodes from the same modality, instead only storing the associations between nodes from different modalities. Consequently, during the first-layer graph convolution, each node only collects affective cues from different modal nodes. For instance, in Fig.~\ref{fig:example_power_1}, the node $n_1^t$ affiliated with the textual modality is exclusively connected to visual nodes (e.g., $n_1^v$, $n_2^v$), thereby assimilating cues from a distinct modality; reciprocally, the visual node $n_1^v$ seizes affective cues solely from textual counterparts.

For the second-layer graph convolution, integrating Eq.~\ref{eq:sgclayer} and Eq.~\ref{eq:sgc1}, the message propagation process can be mathematically formalized as: 
\begin{equation}
\label{eq:sgc2}
\begin{split}
X^{(2)} &= A X^{(1)} \Theta^{(2)} \\
&= A (A X^{(0)} \Theta^{(1)}) \Theta^{(2)} \\
&= A^2 X^{(0)} \Theta^{(1)} \Theta^{(2)}.
\end{split}
\end{equation}
Here, $X^{(2)}$ and $\Theta^{(2)}$ represent the output and learnable parameter for the second layer, respectively. Despite the second layer's features being predicated on those of the first layer, according to Eq.~\ref{eq:sgc2}, the output of the second layer can also be directly computed from the initial input $X^{(0)}$. In our GSF module, since $A^2$ stores the second-hop neighborhood relationships of nodes, the significance of the second-layer graph convolution can be interpreted as nodes assimilating feature information from their second-hop neighbors. Referring to Fig.~\ref{fig:example_power_2}, during the second-layer graph convolution, each node collects affective cues solely from nodes within the same modality. For instance, the textual node $n_1^t$ is connected only to nodes belonging to the textual modality, thereby collecting cues exclusively from intra-modal nodes (e.g., $n_2^t$, $n_3^t$); similarly, the visual node $n_1^v$ gathers affective cues only from visual nodes. Thus, the above message passing process explains why our GraphSmile can still naturally capture intra-modal affective cues without creating intra-modal edges.

The message propagation process for the third-layer graph convolution can be expressed as follows:
\begin{equation}
\label{eq:sgc3}
\begin{split}
X^{(3)} &= A X^{(2)} \Theta^{(3)} \\
&= A (A X^{(1)} \Theta^{(2)}) \Theta^{(3)} \\
&= A (A (A X^{(0)} \Theta^{(1)}) \Theta^{(2)}) \Theta^{(3)} \\
&= A^3 X^{(0)} \Theta^{(1)} \Theta^{(2)} \Theta^{(3)}.
\end{split}
\end{equation}
Here, $X^{(3)}$ signifies the feature matrix resulting from the third layer and $\Theta^{(3)}$ represents the corresponding learnable parameter. Following Eq.~\ref{eq:sgc3}, the third-layer graph convolution, akin to the second layer, can have its output directly computed from the initial features $X^{(0)}$. $A^3$ contains the third-hop neighborhood relationships, thus the implication of the third-layer is that nodes integrate feature information from their third-hop neighbors. As depicted in Fig.~\ref{fig:example_power_3}, during the third-layer graph convolution, each node absorbs affective cues solely from nodes of another modality. For instance, similar to the first-layer convolution, the textual node $n_1^t$ is connected only to nodes of a different modality, i.e., visual nodes, and the visual node $n_1^v$ is connected only to textual nodes.

Through the three examples mentioned above, it can be concluded that the first layer of graph convolution propagates affective cues across different modalities, the second layer of graph convolution conveys affective cues within the same modality, and the third layer of graph convolution once again captures inter-modal associative cues, and so on. Therefore, these examples illustrate that our GSF module can alternately propagate inter-modal and intra-modal affective cues layer by layer, thereby preventing conflicts among multi-source information. Summarizing the patterns from the examples, to capture affective cues from more distant neighbors, we can extend to the $l$-th layer graph convolution:
\begin{equation}
\label{eq:sgcl}
\begin{split}
X^{(l)} &= A X^{(l-1)} \Theta^{(l)} \\
&= A (A X^{(l-2)} \Theta^{(l-1)}) \Theta^{(l)} \\
&= A \cdots (A X^{(0)} \Theta^{(1)}) \cdots \Theta^{(l)} \\
&= (A \cdots A) X^{(0)} (\Theta^{(1)} \cdots \Theta^{(l)}) \\
&= A^l X^{(0)} (\Theta^{(1)} \cdots \Theta^{(l)}).
\end{split}
\end{equation}
Here, $X^{(l)}$ represents the feature matrix at the $l$-th layer, $A^l$ denotes the $l$-th power of the adjacency matrix $A$, which signifies the reach to $l$-hop neighbors, and $\Theta^{(l)}$ are the learnable parameters specific to the $l$-th layer. This formulation allows nodes to aggregate information from increasingly broader neighborhoods as the layer index $l$ increases, thus enriching the representation of each node with more comprehensive affective cues. Via Eq.~\ref{eq:res_connect}, the GSF module is capable of integrating the outputs from all layers. The summation of all layer outputs, i.e., the residual connection, provides two ingenious benefits: (1) it mitigates the over-smoothing problem~\cite{10.1109/TPAMI.2021.3074057} in GNNs; (2) it amalgamates affective cues from multi-hop neighbors.

\subsubsection{Multimodal Integration}\label{sec:multimodal_integration}
The graphs $\mathcal{G}^{tv}$, $\mathcal{G}^{ta}$, and $\mathcal{G}^{va}$ are respectively input into three separate GSF modules for graph convolution operations. After stacking multiple layers of convolutions and combining with Eq.~\ref{eq:res_connect}, three outputs $\mathrm{X}^{tv} $, $ \mathrm{X}^{ta}$, and $\mathrm{X}^{va}$ can be obtained. Here, $\mathrm{X}^{tv}$ is constituted by $\mathrm{X}^{t \leftarrow v}$ and $\mathrm{X}^{v \leftarrow t}$, $\mathrm{X}^{ta}$ is composed of $\mathrm{X}^{t \leftarrow a}$ and $\mathrm{X}^{a \leftarrow t}$, $\mathrm{X}^{va}$ consists of $\mathrm{X}^{v \leftarrow a}$ and $\mathrm{X}^{a \leftarrow v}$. Then, these outputs are summed to obtain a new feature representation:
\begin{equation}
\label{eq:emotion_sum}
\begin{split}
H &= \sigma ((\mathrm{X}^{t \leftarrow v})\Theta_h) + \sigma ((\mathrm{X}^{v \leftarrow t})\Theta_h) \\
&+ \sigma ((\mathrm{X}^{t \leftarrow a})\Theta_h)+ \sigma ((\mathrm{X}^{a \leftarrow t})\Theta_h) \\
&+ \sigma ((\mathrm{X}^{v \leftarrow a})\Theta_h) + \sigma ((\mathrm{X}^{a \leftarrow v})\Theta_h),
\end{split}
\end{equation}
where $\sigma$ represents the activation function, e.g., LeakyReLU, and $\Theta_h$ denotes the shared learnable parameter. The new representation $H$ integrates information from textual-to-visual, textual-to-acoustic, and visual-to-acoustic modalities, providing comprehensive features that encapsulate the multimodal affective cues in dialogues.

\subsection{Emotion Classification}
To achieve emotion classification, $H$ is fed into a classifier that incorporates a linear function followed by a softmax function. The emotion classification is formulated as:
\begin{equation}
\label{eq:emotion_classification}
\begin{split}
\hat{Y} &= \mathtt{softmax}(H\Theta_y + b_y), \\
\hat{e}_i &= \mathtt{argmax}(\hat{y}_i[\tau]),
\end{split}
\end{equation}
where both $\Theta_y$ and $b_y$ are learnable parameters. 
Let $\hat{Y} = [\hat{y}_1, \hat{y}_2, \ldots, \hat{y}_M]$, then $\hat{y}_i$ represents the emotional probability vector of utterance $u_i$, and $\hat{e}_i$ denotes the corresponding predicted emotional label. During the training phase, we employ the categorical cross-entropy loss for network optimization, which can be mathematically expressed as:
\begin{equation}
\label{eq:emotion_loss}
\mathcal{L}_{e} = - \frac {1}{M} \sum_{i=1}^{M}\sum_{j=1}^{C_e} y_{i, j} \log \hat{y}_{i, j},
\end{equation}
where $M$ represents the number of utterances in the dialogue, $C_e$ denotes the number of emotional categories; $y_{i,j}$ signifies the ground-truth probability distribution of the $i$-th utterance $u_i$ in the $j$-th category, and $\hat{y}_{i,j}$ is the predicted probability of utterance $u_i$.

\subsection{Sentiment Dynamics Perception Module}
In dialogue systems, sentiment shift refers to the phenomenon where the sentimental state of two utterances transitions from one category to another. Capturing the dynamic change in sentiments is pivotal for comprehending the sentimental trajectory of a conversation. However, prevailing MERC models predominantly concentrate on contextual modeling of the utterance, exhibiting a dearth of acuity in detecting inter-utterance sentimental changes. This limitation can lead to inaccuracies in the classification of affective states during scenarios characterized by sentiment shifts. To mitigate this shortfall, we introduce an auxiliary task known as the sentiment dynamics perception (SDP) module. Our SDP module serves the following purposes: (1) it models the dynamics of sentimental changes, endowing the model with the capability to not only discern static affective states but also track dynamic sentimental shifts; (2) by pulling utterances with the same sentimental state closer and pushing those with different sentimental states further apart, it aids in enhancing the model's ability to discern between various affective categories; (3) integrated as a complementary task, its loss function can be conjointly optimized with that of the main task, guiding the model to accord significance to the phenomenon of sentiment shifts.

\subsubsection{Shift Label}
Based on the shifts of sentimental states between utterances in a dialogue, we generate a set of sentiment shift labels for that conversation. In a dialogue, if two utterances possess distinct sentiments, i.e., their states transition from one sentiment to another, then the shift label is set to ``1", signifying that a sentiment shift has occurred between them. Conversely, if two utterances share the same sentiment, i.e., there is no alteration in their sentimental state, then the shift label is set to ``0", indicating that no sentiment shift has transpired between them. The annotating process of the shift label $o_{(i,j)}$ can be formulated as follows: 
\begin{equation}
\label{eq:shift_label}
o_{(i,j)} = 
\begin{cases}
1, &\text{if $s_i \neq s_j $}, \\
0, &\text{if $s_i = s_j $}.
\end{cases}
\end{equation}
Here, $s_i$ denotes the sentimental state of the $i$-th utterance $u_i$. It is worth noting that the IEMOCAP datasets do not provide pre-defined sentiment labels. To make them compatible with GraphSmile, we merge their emotion labels, with the specific merging scheme as depicted in Table~\ref{tab:merging}. Additionally, in the CMU-MOSEI dataset, an utterance may correspond to multiple emotion labels, and this annotation way is not suitable for our model. Consequently, we adopt seven sentiments, ranging from highly negative (-3) to highly positive (+3), to substitute the emotion labels for classification.
\begin{table*}[htbp]
\centering
\setlength{\tabcolsep}{10pt}
\caption{\label{tab:merging}Merging Scheme of Emotion Labels}
% \begin{threeparttable}
\begin{tabular}{ccccc}
\toprule
Sentiment &IEMOCAP-6 &IEMOCAP-4 &MELD &CMU-MOSEI \\
\midrule
Negative &Sad, Angry, Frustrated &Sad, Angry &Negative &Highly Negative, Negative, Weakly Negative \\
Neutral &Neutral &Neutral &Neutral &Neutral \\
Positive &Happy, Excited &Happy	&Positive &Weakly Positive, Positive, Highly positive \\
\bottomrule
\end{tabular}
% The MELD dataset itself contains sentiment labels.
% \end{threeparttable}
\end{table*}

\subsubsection{Shift Feature}
In Subsection~\ref{sec:multimodal_integration}, a fused multimodal feature representation $H \in \mathbb{R}^{M \times D}$ is obtained. Here, $M$ denotes the number of utterances in the dialogue, $D$ signifies the feature dimension, and $H$ can be indicated as $[h_1, h_2, \ldots, h_M]$. Based on the feature representation $H$, we construct a set of sentiment shift features. Specifically, given a dialogue containing $M$ utterances, we concatenate the features of these utterances pairwise. Through the aforementioned operation, $M^2$ samples of sentiment shifts can be generated to construct a shift feature representation $T$ with a size of $M^2 \times 2D$.

\begin{figure}[htbp]
\centering
\includegraphics[width=\columnwidth]{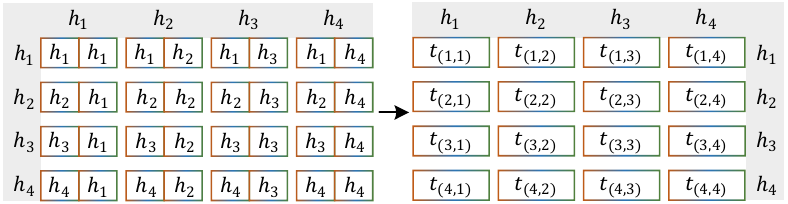}
\caption{Example of generating shift features. Here, $h_i$ denotes the feature vector of the utterance $u_i$.}
\label{fig:shift_feature}
\end{figure}

Fig.~\ref{fig:shift_feature} provides an example of generating shift features. Assuming there are four utterances in the dialogue, their features are represented as $[h_1, h_2, h_3, h_4]$, where $h_i \in \mathbb{R}^{D}$. We initially select the feature vector $h_1$ and concatenate it with all utterances (i.e., $h_1$, $h_2$, $h_3$, and $h_4$), thereby obtaining the feature representations for the first row, i.e., $t_{(1,1)}, t_{(1,2)}, t_{(1,3)}, t_{(1,4)}$. Subsequently, we take $h_2, h_3$, and $h_4$ in sequence and repeat the above procedure. Ultimately, this results in 16 shift samples, with the dimension of the features $t_{(i,j)}$ for each sample being $2D$, where $i=j=1,2,3,4$.

When the number of utterances in a dialogue (i.e., dialogue length) is excessive, recognizing sentiment shifts between an utterance and long distant utterances becomes meaningless. In other words, intuitively, exploring sentiment shifts between nearby utterances is more meaningful. Moreover, for long dialogues, detecting sentiment shifts between all utterances may consume a substantial amount of computational resources. Therefore, we first evenly divide a dialogue into $M/B$ segments ($B$ is the number of utterances per segment), and then generate sentiment shift labels and corresponding features in each segment.

\subsubsection{Shift Classification}
Similar to the emotion classification, we subject the shift feature representation $T$ to a classifier for shift classification, mathematically formulated as:
\begin{equation}
\label{eq:shift_classification}
\begin{split}
\hat{Z} &= \mathtt{softmax}(T \Theta_z + b_z), \\
\hat{o}_{(i,j)} &= \mathtt{argmax}(\hat{z}_{(i,j)}[\eta]),
\end{split}
\end{equation}
where $\Theta_z$ and $b_z$ are learnable parameters for the sentiment shift classifier; $\hat{Z}$ can be denoted as $[\hat{z}_{(1,1)}, \hat{z}_{(1,2)}, \ldots, \hat{z}_{(M,M)}]$, $\hat{z}_{(i,j)}$ represents the shift probability vector constructed based on utterances $u_i$ and $u_j$, and $\hat{o}_{i,j}$ denotes the corresponding predicted shift label. During the training phase, the cross-entropy loss is employed for network learning:
\begin{equation}
\label{eq:shift_loss}
\mathcal{L}_{o} = - \frac {1}{M^2} \sum_{i=1}^{M} \sum_{j=1}^{M} \sum_{k=1}^{C_o} z_{(i, j), k} \log \hat{z}_{(i, j), k},
\end{equation}
where $C_o$ denotes the number of shift categories ($C_o=2$ in this work); $z_{(i,j),k}$ signifies the ground-truth probability distribution of the shift sample constructed based on $u_i$ and $u_j$ in the $k$-th category, and $\hat{z}_{(i, j), k}$ is the predicted probability of the shift sample.

\subsection{Sentiment Classification}
Following the emotion classification, we can easily apply GraphSmile to the MSAC task by adding a sentiment classifier. 
The sentiment classification is represented by the following mathematical formula:
\begin{equation}
\label{eq:sentiment_classification}
\begin{split}
\hat{W} &= \mathtt{softmax}(H \Theta_w + b_w), \\
\hat{s}_i &= \mathtt{argmax}(\hat{w}_i[\phi]).
\end{split}
\end{equation}
Here, $\Theta_w$ and $b_w$ are the learnable parameters specific to the sentiment classifier; $\hat{W}$ can be indicated as $[\hat{w}_1, \hat{w}_2, \ldots, \hat{w}_M]$, and $\hat{w}_i$ represents the sentimental probability vector for utterance $u_i$, with $\hat{s}_i$ denoting the corresponding predicted sentimental label. The loss function for sentiment classification can be expressed as:
\begin{equation}
\label{eq:sentiment_loss}
\mathcal{L}_{s} = - \frac {1}{M} \sum_{i=1}^{M}\sum_{j=1}^{C_s} w_{i, j} \log \hat{w}_{i, j},
\end{equation}
where $C_s$ is the number of sentiment categories, $w_{i,j}$ is the ground-truth probability distribution of utterance $u_i$ in the $j$-th category, and $\hat{w}_{i, j}$ denotes the predicted probability of the utterance. 

\subsection{Total Training Objective}
The MERC task is jointly learned alongside the MSAC task to simultaneously extract emotional and sentimental cues in the dialogue. Uniting all tasks together, we can derive the total training objective, represented by the following formula:
\begin{equation}
\label{eq:training_objective}
\mathcal{L}_{total} = \mathcal{L}_{e} + \lambda_s \mathcal{L}_{s} +\lambda_o \mathcal{L}_{o} + \beta \Vert  \Theta_{total} \Vert,
\end{equation}
where $\lambda_s$ and $\lambda_o$ are trade-off parameters balancing these three losses; $\beta$ is the L2-regularizer weight, and $\Theta_{total}$ is the set of all learnable parameters.

\section{Experimental Setup}\label{sec:experimental_setup}
\subsection{Baseline}
To substantiate the effectiveness of GraphSmile, we select a multitude of baseline models for comparison. These baselines are categorized into graph-based models and non-graph-based models. The former encompasses COGMEN~\cite{joshi-etal-2022-cogmen}, Joyful~\cite{li-etal-2023-joyful}, DER-GCN~\cite{10458270}, DCGCN~\cite{10418539}, MMGCN~\cite{hu2021mmgcn}, MM-DFN~\cite{hu2022mmdfn}, M3Net~\cite{10203083}; whereas the latter includes MetaDrop~\cite{chen2021learning}, UniMSE~\cite{hu2022unimse}, EmoLR~\cite{10.1162/tacl_a_00614}, DialogueCRN~\cite{hu-etal-2021-dialoguecrn}, SACL-LSTM~\cite{hu-etal-2023-supervised}.

\subsection{Dataset}
We conduct experiments across three datasets, including IEMOCAP, MELD, and CMU-MOSEI. The statistical information of all datasets are shown in Table~\ref{tab:statistics}. 
IEMOCAP~\cite{busso2008iemocap} is a renowned multimodal corpus of dyadic conversations that captures emotional interactions from ten participants in both scripted and improvisational settings. 
We employ two IEMOCAP ways, one including all six emotions, i.e., IEMOCAP-6, and the other including four emotions, i.e., IEMOCAP-4, with Excited and Frustrated removed. 
MELD~\cite{poria2019meld} is an esteemed collection of multimodal multiparty conversations, encompassing 1,433 dialogues from 304 speakers in the television series ``Friends", totaling 13,708 utterances. 
CMU-MOSEI~\cite{bagher-zadeh-etal-2018-multimodal} is an extensive multimodal affective collection that aggregates video commentary segments from YouTube, featuring a wide range of 1,000 speakers. 
To be applicable to GraphSmile, we only use the sentiment annotation in this work. Referring to previous work~\cite{joshi-etal-2022-cogmen}, we divide sentiment intensity into seven distinct classes, encompassing Highly Negative, Negative, Weakly Negative, Neutral, Weakly Positive, Positive, and Highly Positive. The corresponding intensity ranges for these classes are [-3, -2), [-2, -1), [-1, 0), 0, (0, 1], (1, 2], and (2, 3], respectively.
\begin{table}[htbp]
\centering
\setlength{\tabcolsep}{8pt}
\caption{\label{tab:statistics}Statistical Information of All Datasets}
\begin{threeparttable}
\begin{tabular}{ccccccc}
\toprule
\multirow{2}{*}{Dataset} & \multicolumn{2}{c}{Conversation} & \multicolumn{2}{c}{Utterance} \\
\cmidrule(lr){2-3} \cmidrule(lr){4-5}
  & train+val & test & train+val & test \\
\midrule
  IEMOCAP-6 & 120 & 31 & 5,810 & 1,623 \\
  IEMOCAP-4 & 120 & 31 & 3,600 & 943 \\
  MELD & 1,153 & 280 & 11,098 & 2,610 \\
  CMU-MOSEI & 2,549 & 646 & 18,198 & 4,662 \\
\bottomrule
\end{tabular}
The 10\% of dialogues from the combination of training and validation sets are selected as the validation set.
\end{threeparttable}
\end{table}

\subsection{Training Detail}
We utilize accuracy (ACC) and weighted F1 (WF1) as the primary evaluation metrics, with F1 being used as the metric for each individual emotion category. To test the significance of performance improvements, a paired t-test is performed with the default significance level set at 0.05. All experiments are conducted on an NVIDIA GeForce RTX 3090, using CUDA version 11.7 and the deep learning framework PyTorch 2.0.0. AdamW is employed as the optimizer, with the L2 regularization factor of 1e-3. Other hyperparameter settings are shown in Table~\ref{tab:hyperparameter_emo}. The code is released at https://github.com/lijfrank-open/GraphSmile.
\begin{table}[htbp]
\centering
\setlength{\tabcolsep}{5.5pt}
\caption{\label{tab:hyperparameter_emo}Description of Partial Hyperparameters}
\begin{threeparttable}
\begin{tabular}{ccccccccc}
\toprule
Dataset & LR & BS & DR & $L$ & $P$ & $B$ & $\lambda_s$ & $\lambda_o$\\
\midrule
IEMOCAP-6 & 1e-4 & 16 & 0.2 & 7 & 17 & 19 & 1.0 & 0.7 \\
IEMOCAP-4 & 3e-4 & 16 &  0.2  & 4 & 5 & 10 & 0.6 & 0.6 \\
MELD & 7e-5 & 16 & 0.2 & 5 & 3 & 3 & 0.5 & 0.2 \\
CMU-MOSEI & 8e-5 & 32 & 0.4 & 2 & 5 & 2 & 0.8 & 1.0 \\
\bottomrule
\end{tabular}%
The abbreviations ``LR", ``BS", and ``DR" denote the learning rate, batch size, and dropout rate, respectively.
\end{threeparttable}
\end{table}

\section{Experimental Result}\label{sec:experimental_result}
\subsection{Main Result for Emotion Recognition}
\begin{table*}[htbp]
\centering
\small
\setlength{\tabcolsep}{8pt}
\caption{\label{tab:main_result_emo}Main Results for the MERC Task on All Datasets}
\begin{threeparttable}
\begin{tabular}{cccccccccc}
\toprule
\multirow{2}{*}{Method} & \multicolumn{9}{c}{IEMOCAP-6} \\
\cmidrule(lr){2-10} 
& Happy & Sad & Neutral & Angry & Excited & Frustrated & \multicolumn{2}{c}{ACC} & WF1 \\
\midrule
MetaDrop$^\dagger$ & -- & -- & -- & -- & -- & -- & \multicolumn{2}{c}{69.38} & 69.59 \\
UniMSE$^\dagger$ & -- & -- & -- & -- & -- & -- & \multicolumn{2}{c}{70.56} & 70.66 \\
COGMEN$^\dagger$& 51.90 & 81.70 & 68.60 & 66.00 & 75.30 & 58.20 & \multicolumn{2}{c}{68.20} & 67.60 \\
EmoLR$^\dagger$ & -- & -- & -- & -- & -- & -- & \multicolumn{2}{c}{68.53}  & 68.12 \\
Joyful$^\dagger$ & \underline{60.94} & \textbf{84.42} & 68.24 & 69.95 & 73.54 & \underline{67.55} & \multicolumn{2}{c}{70.55} & 71.03 \\
DER-GCN$^\dagger$ & 58.80 & 79.80 & 61.50 & \textbf{72.10} & 73.30 & \textbf{67.80} & \multicolumn{2}{c}{69.70} & 69.40 \\
DCGCN$^\dagger$ & -- & -- & -- & -- & -- & -- & \multicolumn{2}{c}{--} & 68.31 \\
% \midrule
DialogueCRN$^\ddagger$ & 53.85 & 82.66 & 71.03 & 62.33 & \underline{77.64} & 58.81 & \multicolumn{2}{c}{68.70} & 68.82 \\
MMGCN$^\ddagger$ & 47.10 & 81.91 & 66.44 & 63.51 & 76.17 & 59.06 & \multicolumn{2}{c}{67.10} & 66.81 \\
MM-DFN$^\ddagger$ & 43.36 & \underline{83.23} & 70.03 & 70.19 & 73.11 & 64.01 & \multicolumn{2}{c}{69.44} & 68.83 \\
SACL-LSTM$^\ddagger$ & 51.30 & 82.25 & \textbf{71.39} & 67.78 & 75.26 & 66.94 & \multicolumn{2}{c}{70.55} & 70.60 \\
M3Net$^\ddagger$ & 60.93 & 78.84 & 70.14 & 68.06 & 77.11 & 67.42 & \multicolumn{2}{c}{\underline{70.92}} & \underline{71.07} \\
\midrule
GraphSmile & \textbf{63.09} & 83.16 & \underline{71.07} & \underline{71.38} & \textbf{79.66} & 66.84 & \multicolumn{2}{c}{\textbf{72.77}} & \textbf{72.81} \\
\midrule
\midrule
\multirow{2}{*}{Method} & \multicolumn{9}{c}{IEMOCAP-4}\\
\cmidrule(lr){2-10}
  & \multicolumn{2}{c}{Happy} & Sad & \multicolumn{2}{c}{Neutral} & Angry & \multicolumn{2}{c}{ACC} & WF1 \\
\midrule
COGMEN$^\dagger$ & \multicolumn{2}{c}{--} & -- & \multicolumn{2}{c}{--} & -- & \multicolumn{2}{c}{--} & 84.50 \\
Joyful$^\dagger$ & \multicolumn{2}{c}{80.10} & \underline{88.10} & \multicolumn{2}{c}{\underline{85.10}} & \textbf{88.10} & \multicolumn{2}{c}{--} & \underline{85.70} \\
% \midrule
DialogueCRN$^\ddagger$ & \multicolumn{2}{c}{75.64} & 82.05 & \multicolumn{2}{c}{81.16} & 85.22 & \multicolumn{2}{c}{81.34} & 81.28 \\
MMGCN$^\ddagger$ & \multicolumn{2}{c}{79.56} & 79.33 & \multicolumn{2}{c}{79.27} & 81.42 & \multicolumn{2}{c}{79.75} & 79.72 \\
MM-DFN$^\ddagger$ & \multicolumn{2}{c}{\underline{80.30}} & 80.41 & \multicolumn{2}{c}{79.16} & \underline{85.64} & \multicolumn{2}{c}{80.91} & 80.83 \\
SACL-LSTM$^\ddagger$ & \multicolumn{2}{c}{75.64} & 78.75 & \multicolumn{2}{c}{81.83} & 85.47 & \multicolumn{2}{c}{80.70} & 80.74 \\
M3Net$^\ddagger$ & \multicolumn{2}{c}{80.00} & 85.16 & \multicolumn{2}{c}{84.63} & 81.90 & \multicolumn{2}{c}{\underline{83.67}} & 83.57 \\
\midrule
GraphSmile & \multicolumn{2}{c}{\textbf{87.94}} & \textbf{88.67} & \multicolumn{2}{c}{\textbf{85.41}} & 84.73 & \multicolumn{2}{c}{\textbf{86.53}} & \textbf{86.52} \\
\midrule
\midrule
\multirow{2}{*}{Method}  & \multicolumn{9}{c}{MELD} \\
\cmidrule(lr){2-10}
& Neutral & Surprise & Fear & Sadness & Joy & Disgust & Anger & ACC & WF1 \\
\midrule
MetaDrop$^\dagger$ & -- & -- & -- & -- & -- & -- & -- & 66.63 & \underline{66.30} \\
UniMSE$^\dagger$  & -- & -- & -- & -- & -- & -- & -- & 65.09  & 65.51 \\
EmoLR$^\dagger$ & -- & -- & -- & -- & -- & -- & -- &66.69  & 65.16 \\
Joyful$^\dagger$ & 76.80 & 51.91 & -- & 41.78 & 56.89 & -- & 50.71 & 62.53 & 61.77 \\
DER-GCN$^\dagger$ & \textbf{80.60} & 51.00 & 10.40 & 41.50 & 64.30 & 10.30 & \textbf{57.40} & 66.80  & 66.10 \\
DCGCN$^\dagger$ & -- & -- & -- & -- & -- & -- & -- & --  & 66.25 \\
% \midrule
DialogueCRN$^\ddagger$ & 76.15 & 56.72 & \underline{18.67} & 38.29 & 63.21 & 27.69 & 50.67 & 62.38 & 63.32 \\
MMGCN$^\ddagger$ & 78.62 & 57.78 & 3.77 & 40.35 & 63.60 & 12.20 & \underline{53.68} & 66.02 & 64.55 \\
MM-DFN$^\ddagger$ & 79.84 & 58.43 & 15.79 & 31.65 & 64.01 & 28.04 & 53.60 & \underline{67.05} & 65.21 \\
SACL-LSTM$^\ddagger$ & 77.42 & 58.50 & \textbf{20.41} & 39.58 & 62.76 & \textbf{34.71} & 52.08 & 64.52 & 64.55 \\
M3Net$^\ddagger$ & 79.14 & \textbf{59.54} & 13.33 & \textbf{42.86} & \textbf{65.05} & 21.69 & 53.54 & 66.59 & 65.83 \\
\midrule
GraphSmile & \underline{80.35} & \underline{59.11} & 18.18 & \underline{42.46} & \underline{64.99} & \underline{32.43} & 53.67 & \textbf{67.70} & \textbf{66.71} \\
\midrule
\midrule
\multirow{2}{*}{Method}  & \multicolumn{9}{c}{CMU-MOSEI} \\
\cmidrule(lr){2-10}
& HNegative & Negative & WNegative & Neutral & WPositive & Positive & HPositive & ACC & WF1 \\
\midrule
COGMEN$^\dagger$ & -- & -- & -- & -- & --  & -- & -- & 43.90 & -- \\
% \midrule
DialogueCRN$^\ddagger$ & 0.00 & 4.29 & 7.98 & 25.09 & 51.80 & 3.22 & 0.00 & 37.88 & 26.55 \\
MMGCN$^\ddagger$ & 0.00 & \underline{19.51} & \textbf{43.75} & \textbf{42.45} & 54.64 & \textbf{36.13} & 0.00 & \underline{45.67} & \underline{44.11} \\
MM-DFN$^\ddagger$ & 0.00 & 16.98 & 37.94 & \underline{39.64} & \underline{56.58} & 32.51 & \textbf{8.51} & 45.29 & 42.98 \\
SACL-LSTM$^\ddagger$ & 0.00 & 0.00 & 0.00 & 17.87 & 55.28 & 0.00 & 0.00 & 38.60 & 25.95 \\
M3Net$^\ddagger$ & 0.00 & 12.50 & 37.26 & 33.29 & 56.10 & 33.94 & 0.00 & 43.67 & 41.12 \\
\midrule
GraphSmile & 0.00 & \textbf{28.79} & \underline{43.50} & 39.19 & \textbf{57.30} & \underline{35.95} & \underline{6.25} & \textbf{46.82} & \textbf{44.93} \\
\bottomrule
\end{tabular}
The marker $^\dagger$ indicates the results (\%) from the original paper, and the marker $^\ddagger$ indicates the replication results (\%) based on the open source code and our dataset setting. In DialogueCRN and SACL-LSTM, multimodal features are directly concatenated to meet their input requirements. The abbreviations ``HNegative'' and ``WNegative'' denote Highly Negative and Weakly Negative, respectively, and other sentiments by analogy.
\end{threeparttable}
\end{table*}
The results for the MERC task on all datasets are reported in Table~\ref{tab:main_result_emo}. Overall, our proposed GraphSmile outperforms all baseline models on these datasets. Among the reproduced baseline models, M3Net achieves relatively balanced results across all datasets, benefiting from GNNs' thorough exploration of multivariate relationships and multi-frequency signals. The poor performance of DialogueCRN and SACL-LSTM on the CMU-MOSEI dataset indicates that models based on recurrent neural networks struggle to capture the emotional cues present in that dataset. GraphSmile surpasses all graph-based methods on the IEMOCAP (6-way) dataset. For instance, GraphSmile's accuracy is 5.67\%, 3.33\%, and 1.85\% higher than that of MMGCN, MM-DFN, and M3Net on this dataset, respectively. On the IEMOCAP (4-way) dataset, our model improves the weighted F1 score by 2.02\% and 0.82\% compared to the strong baselines COGMEN and Joyful, respectively. These advanced results are attributed to the significant role played by the GSF module in cross-modal modeling. On the other two more challenging datasets, GraphSmile continues to maintain its advantage. For example, compared to M3Net, GraphSmile's accuracy is 1.11\% and 3.15\% higher on the MELD and CMU-MOSEI datasets, respectively. In addition to the contribution of the GSF module, the contribution of SDP in sentimental dynamics modeling also contributes to the performance advantage of GraphSmile.

The table also presents the F1 scores for each emotion. In the IEMOCAP (6-way) dataset, compared to other emotions, all models achieve the highest and second-highest scores on Sad and Excited, respectively, and the lowest on Happy. This indicates that Sad and Excited are relatively easier to distinguish than Happy. Relative to other models, our GraphSmile demonstrates a more balanced performance across all emotions, especially achieving the best score on the challenging emotion Happy. As can be seen from the results of the IEMOCAP (4-way) dataset, due to the reduction in the number of categories in the dataset, these models have achieved satisfactory results on all emotions. Among these baselines (with the exception of MMGCN), there is always one or more emotions with significantly lower scores than the others, indicating performance imbalance. In contrast, GraphSmile consistently achieves relatively high results across all emotions. Similar to the results on the IEMOCAP (6-way) dataset, our GraphSmile scores significantly higher on Happy than other models. In the MELD dataset, compared to other emotions, all models perform poorly on Fear and Disgust. The direct cause of this phenomenon is that these two emotions are extreme minority classes in the MELD dataset. Since samples labeled as Neutral constitute the majority of all samples, all models achieve the best F1 scores on Neutral. Compared to other models, SACL-LSTM and GraphSmile achieves the best and second-best scores on the minority class Disgust, respectively. In the CMU-MOSEI dataset, as each utterance may correspond to multiple emotions, we employ seven sentiment labels instead of emotion ones for our experiments. Table~\ref{tab:main_result_emo} shows that the classification task for the CMU-MOSEI dataset is quite challenging, with all models struggling to distinguish each category, particularly failing to correctly identify Highly Negative. Compared to the baselines, GraphSmile achieves the best results on Negative, Weakly Positive, and Highly Positive.

\subsection{Main Result for Sentiment Analysis}
\begin{table*}[htbp]
\centering
\small
\setlength{\tabcolsep}{9pt}
\caption{\label{tab:main_result_sen}Main Results for the MSAC Task on All Datasets}
\begin{threeparttable}
\begin{tabular}{ccccccccccc}
\toprule
\multirow{2}{*}{Method} & \multicolumn{5}{c}{IEMOCAP-6} & \multicolumn{5}{c}{IEMOCAP-4} \\
\cmidrule(lr){2-6} \cmidrule(lr){7-11}
& Negative & Neutral & Positive & ACC & WF1 & Negative & Neutral & Positive & ACC & WF1 \\
\midrule
DialogueCRN$^\ddagger$ & 87.87 & 67.09 & \underline{89.96} & 83.43 & 83.52 & 83.94 & 76.81 & 79.39 & 80.70 & 80.34 \\
MMGCN$^\ddagger$ & 87.07 & 68.17 & 88.31 & 82.75 & 82.94 & 84.97 & 78.80 & 79.09 & 81.87 & 81.56 \\
MM-DFN$^\ddagger$ & \textbf{89.08} & 71.92 & 87.85 & 84.60 & 84.69 & 84.79 & 76.99 & 80.95 & 81.34 & 81.03 \\
SACL-LSTM$^\ddagger$ & \underline{88.82} & \textbf{72.95} & 89.85 & \textbf{85.15} & \textbf{85.35} & 85.23 & 80.23 & 80.53 & 82.61 & 82.47 \\
M3Net$^\ddagger$ & 87.52 & 68.35 & \textbf{91.93} & 84.29 & 84.19 & 87.14 & 82.61 & 81.53 & 84.52 & 84.44 \\
\midrule
GraphSmile$^\nmid$ & 86.87 & 69.06 & 89.21 & 83.73 & 83.66 & \underline{87.76} & \underline{82.93} & \textbf{87.23} & \underline{85.79} & \underline{85.71} \\
GraphSmile$^\nparallel$ & 88.28 & \underline{72.89} & 89.93 & \underline{84.97} & \underline{85.09} & \textbf{89.45} & \textbf{85.53} & \underline{87.02} & \textbf{87.49} & \textbf{87.48} \\
\midrule
\midrule
\multirow{2}{*}{Method} & \multicolumn{5}{c}{MELD} & \multicolumn{5}{c}{CMU-MOSEI}\\
\cmidrule(lr){2-6} \cmidrule(lr){7-11}
& Neutral & Positive & Negative & ACC & WF1 & Negative & Neutral & Positive & ACC & WF1 \\
\midrule
DialogueCRN$^\ddagger$ & \textbf{80.05} & 66.67 & 69.34 & \underline{74.18} & \underline{73.96} & 20.87 & 12.71 & 70.51 & 54.11 & 48.50 \\
MMGCN$^\ddagger$ & 79.03 & 66.60 & 69.36 & 73.56 & 73.46 & \textbf{58.10} & 38.62 & 78.47 & 66.66 & \underline{65.89} \\
MM-DFN$^\ddagger$ & 77.89 & 66.05 & \textbf{70.83} & 73.30 & 73.27 & 53.67 & 36.11 & \underline{78.67} & 66.35 & 64.62 \\
SACL-LSTM$^\ddagger$ & 79.08 & 66.67 & 69.16 & 73.56 & 73.44 & 10.59 & 0.00 & 71.73 & 55.28 & 44.50 \\
M3Net$^\ddagger$ & 78.46 & \underline{67.12} & 69.44 & 73.33 & 73.32 & 53.08 & 25.92 & 77.64 & 64.72 & 61.66 \\
\midrule
GraphSmile$^\nmid$ & 79.15 & 64.84 & 70.08 & 73.60 & 73.40 & 56.14 & \underline{39.80} & 78.18 & \underline{66.83} & 65.69 \\
GraphSmile$^\nparallel$ & \underline{79.35} & \textbf{67.89} & \underline{70.72} & \textbf{74.44} & \textbf{74.31} & \underline{56.99} & \textbf{41.02} & \textbf{79.36} & \textbf{67.73} & \textbf{66.73} \\
\bottomrule
\end{tabular}
The marker $^\ddagger$ indicates the replication results (\%) based on the open source code and our dataset setting. The marker $^\nmid$ indicates the result (\%) of the MSAC task with the hyperparameters held constant (i.e., the settings in Table~\ref{tab:hyperparameter_emo}), while the marker $^\nparallel$ represents the result (\%) after optimizing the hyperparameters.
\end{threeparttable}
\end{table*}
The proposed GraphSmile not only demonstrates exceptional performance in MERC but also successfully extends to the MSAC task. In Table~\ref{tab:main_result_sen}, we collect the experimental results of GraphSmile performing the MSAC task across multiple datasets. The results obtained based on the hyperparameter settings in Table~\ref{tab:main_result_emo} show that even without task-specific hyperparameter tuning, GraphSmile can still achieve commendable performance, especially surpassing all baselines on the IEMOCAP (4-way) dataset. The reduction in the number of categories leads to a significant performance improvement for all models on datasets other than IEMOCAP (4-way). After hyperparameter optimization specifically for the MSAC task, GraphSmile's performance is further enhanced, outperforming all baselines on most datasets. Although SACL-LSTM achieves the best scores on the IEMOCAP (6-way) dataset, its performance on other datasets is not prominent. Similar to the results of the MERC task, DialogueCRN and SACL-LSTM show relatively weaker capabilities in capturing sentimental cues on the CMU-MOSEI dataset, with their performance significantly lower than other models. Additionally, the table lists the F1 score for each sentiment category in detail. In the IEMOCAP (6-way) dataset, the model scores lower in identifying Neutral, indicating that distinguishing Neutral is more challenging compared to other sentiments. A similar phenomenon is observed in the CMU-MOSEI dataset. In the IEMOCAP (4-way) dataset, the recognition of Negative is the most effective, while in the MELD dataset, the recognition score of Neutral is the highest. These results indicate that there are variations in the recognition difficulty for different datasets and sentiments.

\subsection{Ablation Study}
\begin{table*}[htbp]
\centering
\small
\setlength{\tabcolsep}{16pt}
\caption{\label{tab:ablation_study}Results for Ablation Studies on All Datasets}
\begin{threeparttable}
\begin{tabular}{cccccccccc}
\toprule
\multirow{2}{*}{Method}  & \multicolumn{2}{c}{IEMOCAP-6} & \multicolumn{2}{c}{IEMOCAP-4} & \multicolumn{2}{c}{MELD} & \multicolumn{2}{c}{CMU-MOSEI} \\
\cmidrule(lr){2-3} \cmidrule(lr){4-5} \cmidrule(lr){6-7} \cmidrule(lr){8-9}
& ACC & WF1 & ACC & WF1 & ACC & WF1 & ACC & WF1 \\
\midrule
GraphSmile$^\sharp$ & 72.77 & 72.81 & 86.53 & 86.52 & 67.70 & 66.71 & 46.82 & 44.93 \\
\midrule
No $\mathcal{L}_s$$^\sharp$ & 71.53 & 71.52 & 85.15 & 85.15 & 67.39 & 66.30 & 45.08 & 43.37 \\
No $\mathcal{L}_o$$^\sharp$ & 71.41 & 71.42 & 84.52 & 84.50 & 67.62 & 66.59 & 45.20 & 44.03 \\
No $\mathcal{L}_s$ \& $\mathcal{L}_o$$^\sharp$ & 70.67 & 70.49 & 84.41 & 84.40 & 67.55 & 66.43 & 44.29 & 43.54 \\
\midrule
No Res$^\sharp$ & 61.98 & 61.35 & 79.64 & 79.58 & 52.07 & 46.24 & 41.75 & 40.12 \\
No FC-Res$^\sharp$ & 70.43 & 70.39 & 85.37 & 85.36 & 67.36 & 66.30 & 45.03 & 43.28 \\
No Seg$^\sharp$ & 71.66 & 71.69 & 85.47 & 85.43 & 67.59 & 66.59 & 45.89 & 44.53 \\
\midrule
No Textual$^\sharp$ & 55.70 & 54.23 & 62.88 & 62.22 & 50.15 & 42.27 & 32.98 & 28.28 \\
No Visual$^\sharp$ & 71.29 & 71.23 & 86.21 & 86.20 & 67.39 & 66.26 & 45.32 & 43.57 \\
No Acoustic$^\sharp$ & 68.45 & 68.26 & 83.35 & 83.28 & 67.55 & 66.40 & 46.56 & 44.57 \\
\midrule
\midrule
GraphSmile$^\natural$ & 84.97 & 85.09 & 87.49 & 87.48 & 74.44 & 74.31 & 67.73 & 66.73 \\
\midrule
No $\mathcal{L}_e$$^\natural$ & 84.41 & 84.53 & 85.47 & 85.40 & 73.98 & 73.84 & 66.59 & 65.91 \\
No $\mathcal{L}_o$$^\natural$ & 84.10 & 84.18 & 85.26 & 85.23 & 74.18 & 74.10 & 66.71 & 66.20 \\
No $\mathcal{L}_e$ \& $\mathcal{L}_o$$^\natural$ & 83.30 & 83.54 & 85.58 & 85.55 & 73.91 & 73.77 & 66.54 & 65.64 \\
\midrule
No Res$^\natural$ & 78.56 & 78.25 & 81.02 & 80.83 & 58.08 & 56.61 & 62.26 & 61.06 \\
No FC-Res$^\natural$ & 84.35 & 84.38 & 86.00 & 85.90 & 73.72 & 73.64 & 66.20 & 65.69 \\
No Seg$^\natural$ & 83.49 & 83.64 & 85.37 & 85.32 & 74.37 & 74.22 & 66.44 & 65.88 \\
\midrule
No Textual$^\natural$ & 71.04 & 70.42 & 66.28 & 60.76 & 52.72 & 52.10 & 54.95 & 47.80 \\
No Visual$^\natural$ & 84.53 & 84.69 & 85.58 & 85.57 & 73.83 & 73.75 & 66.30 & 65.81 \\
No Acoustic$^\natural$ & 83.24 & 82.87 & 84.31 & 84.25 & 73.72 & 73.62 & 66.47 & 65.57 \\
\bottomrule
\end{tabular}
The marker $^\sharp$ indicates the result (\%) of the MERC task, while the marker $^\natural$ indicates the result (\%) of the MSAC task. The abbreviation ``Res" refers to our designed residual connection, ``FC-Res" denotes the fully connected layer in the residual connection, and ``Seg" signifies the utterance segmentation strategy when executing shift classification.
\end{threeparttable}
\end{table*}
To comprehensively evaluate the functionality of various components in GraphSmile, we report a series of ablation results in Table~\ref{tab:ablation_study}. These studies systematically examine several key factors, including multitask learning, residual connection, utterance segmentation, and modality configuration.

\textit{Multitask Learning}: In ablation studies of multitask learning, we design three experiments, i.e., removing the sentiment classification task (No $\mathcal{L}_s$), removing the shift classification task (No $\mathcal{L}_o$), and removing both tasks simultaneously (No $\mathcal{L}_s$ \& $\mathcal{L}_o$). The results indicate that the removal of any single task adversely affects the overall performance of the model. In the MERC task, the performance degradation attributable to the excision of $\mathcal{L}_s$ substantiates the capacity of the MSAC task to enhance the MERC task's comprehension of emotions in a dialogue through knowledge complementation. Reciprocally, in the MSAC task, the performance decline induced by the ablation of $\mathcal{L}_e$ evidences that the MERC task can facilitate the MSAC task in understanding sentiments. In both tasks, the decrement in efficacy consequent to the elimination of $\mathcal{L}_o$ corroborates our SDP module can enable the model to capture the sentiment dynamics in a conversation, thus ensuring precise classification in scenarios with abrupt sentimental transitions. In addition, on the IEMOCAP dataset, the removal of $\mathcal{L}_o$ leads to a more significant performance drop compared to the removal of $\mathcal{L}_s$; conversely, on the MELD and CMU-MOSEI datasets, the observed phenomenon is the opposite of that on the IEMOCAP. This could be attributed to the abundance of long dialogues contained in the IEMOCAP dataset, whose sentimental changes is more complex relative to the other two datasets, thus necessitating a more urgent need for modeling sentiment dynamics.

\textit{Residual Connection}: During evaluating the influence of residual connections on the efficacy, it is observed that the removal of residual connections (No Res) markedly impairs the model's performance across all datasets. These outcomes can be ascribed to several pivotal factors: (1) Compromised modality integration capability. The elimination of residual connections precipitates a direct diminution in the model's capacity to integrate intra-modal contextual and inter-modal associative cues that are propagated in an alternating fashion. (2) Loss of neighboring information. In the absence of residual connections, the model may exhibit a propensity to prioritize long-range information propagation, forgetting the affective cues proffered by neighboring nodes. (3) Over-smoothing predicament. The model devoid of residual connections are susceptible to the over-smoothing dilemma as the network's depth intensifies, leading to a convergence in the feature representations of disparate nodes. (4) Constrained affective expression facility. Simplified graph convolutional layers exhibit constraints in expressing emotions and sentiments due to the lack of non-linear transformation, a deficiency that is amplified by the absence of residual connections. To delve deeper into the role of the fully connected layer in the residual connection, an ablation experiment is conducted by selectively removing the fully connected components of residual connections (No FC-Res). The experimental outcomes indicate a decrement in performance relative to the complete GraphSmile across the four datasets, thereby underscoring the significant role of the fully connected layer in augmenting the model's capacity for affective expression. 

\textit{Utterance Segmentation}: In the shift classification task, we adopt a segmentation strategy for utterances. The relinquishment of the segmentation strategy (No Seg) results in a degradation in the model's performance. This substantiates the efficacy of the strategy in capturing sentiment dynamics, i.e., facilitating the model's perception of sentimental transitions and continuities among proximate utterances, thereby enabling a more nuanced comprehension of useful sentimental shifts and continuities by the model.

\textit{Modality Configuration}: The empirical outcomes in modality ablation underscores a degradation in the model's holistic performance following the removal of any modality, thereby revealing the indispensable complementarity of multimodal data in the MERC task. Specifically, across all examined datasets, the elimination of the textual modality (No Textual) engenders a markedly pronounced decrement in performance relative to other modalities. This revelation accentuates the dominant role of the textual modality in the MERC task, proffering more critical emotional information. Concurrently, it implies that the visual and acoustic modalities may carry relatively less emotional cues. Such disparities may stem from the noise embedded in visual and acoustic modalities, such as irrelevant background information, ambient sounds, and multiple facial expressions. Further analysis of the experimental results from the IEMOCAP dataset evinces that the exclusion of the visual modality (No Visual) exerts a relatively muted influence on the model's efficacy compared to the excision of the acoustic modality (No Acoustic). This observation posits that in the IEMOCAP dataset, the visual modality contributes less to MERC than the other two modalities. Based on the results from the MELD and CMU-MOSEI datasets, it can be concluded that the contributions of the visual and acoustic modalities to the model's performance on these two datasets are comparable, yet both are inferior to the textual modality.

\subsection{Time and Memory Overhead}
\begin{table*}[htbp]
\centering
\setlength{\tabcolsep}{12.5pt}
\caption{\label{tab:time_memory}Time and Memory Overhead for the MERC Task on All Datasets}
\begin{threeparttable}
\begin{tabular}{ccccccccc}
\toprule
\multirow{2}{*}{Method} & \multicolumn{2}{c}{IEMOCAP-6} & \multicolumn{2}{c}{IEMOCAP-4} & \multicolumn{2}{c}{MELD} & \multicolumn{2}{c}{CMU-MOSEI} \\
\cmidrule(lr){2-3} \cmidrule(lr){4-5} \cmidrule(lr){6-7} \cmidrule(lr){8-9}
& Time & Memory & Time & Memory & Time & Memory & Time & Memory \\
\midrule
DialogueCRN$^\ddagger$ & 5.82 & 2411.95 & 4.13 & 992.86 & 15.58 & 557.15 & 37.26 & 766.55 \\
MMGCN$^\ddagger$ & 1.81 & 547.13 & 1.59 & 251.09 & 13.97 & 191.24 & 29.23 & 344.52 \\
MM-DFN$^\ddagger$ & 2.46 & 1939.07 & 2.03 & 909.80 & 22.87 & 767.48 & 37.73 & 680.09 \\
SACL-LSTM$^\ddagger$ & 3.06 & 3353.43 & 2.48 & 1257.00 & 14.20 & 2042.30 & 14.76 & 1634.85 \\
M3Net$^\ddagger$ & 6.58 & 12841.27 & 2.18 & 3782.62 & 4.70 & 761.19 & 11.12 & 4549.76 \\
\midrule
GraphSmile & 1.12 & 1840.89 & 0.67 & 509.88 & 5.21 & 513.33 & 7.38 & 467.13 \\
\bottomrule
\end{tabular}
The marker $^\ddagger$ indicates the replication results based on the open source code and our dataset setting. The term ``Time" refers to the duration (s) required for training and testing in each epoch, and ``Memory" refers to the consumption (MB) of allocated and reserved memory.
\end{threeparttable}
\end{table*}
In Table~\ref{tab:time_memory}, we report the running time and memory usage of our model and its baselines for the MERC task across all datasets. Note that all these experiments are conducted on a single NVIDIA GeForce RTX 3090. Relative to other models, MMGCNN and GraphSmile exhibit an absolute advantage in time expenditure on the IEMOCAP dataset, yet MMGCN's time usage on the other two datasets is significantly higher than that of GraphSmile. M3Net and GraphSmile achieve lower time consumption on the MELD dataset, while M3Net's running time on other datasets is inferior to GraphSmile. The running times of DialogueCRN, MM-DFN, and SACL-LSTM across all datasets are unsatisfactory. MMGCN's memory consumption is considerably lower than other models on all datasets, but its overly simplistic model leads to poor performance on most datasets. M3Net and SACL-LSTM have much higher memory consumption on multiple datasets compared to other models, and the memory expenditure of DialogueCRN and MM-DFN is also higher than that of the proposed model. In summary, GraphSmile not only achieves the best accuracy and F1 scores but also balances the costs of time and memory.

\subsection{Impact of Network Depth}
\begin{figure*}[htbp]
\centering
{\includegraphics[width=0.45\linewidth]{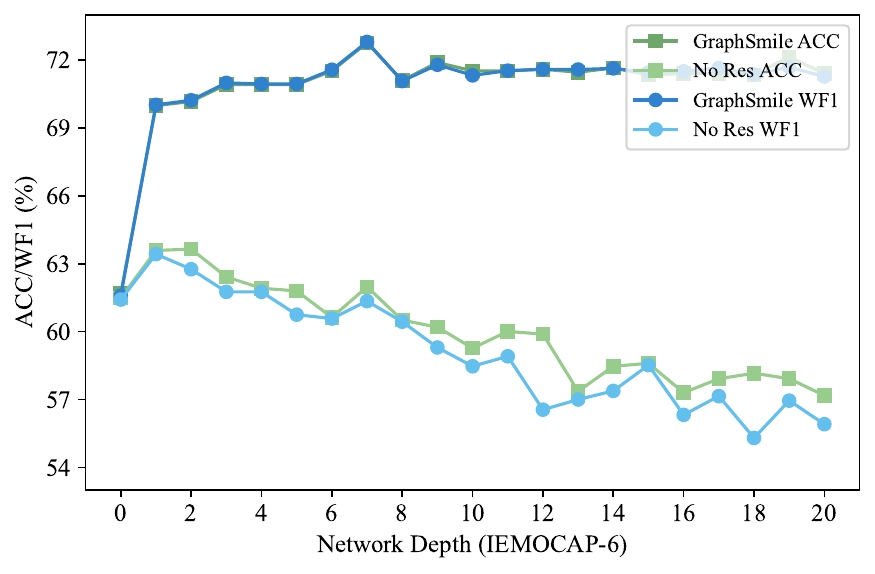}%
\label{fig:network_depth_iemocap6}}
\hfil
{\includegraphics[width=0.45\linewidth]{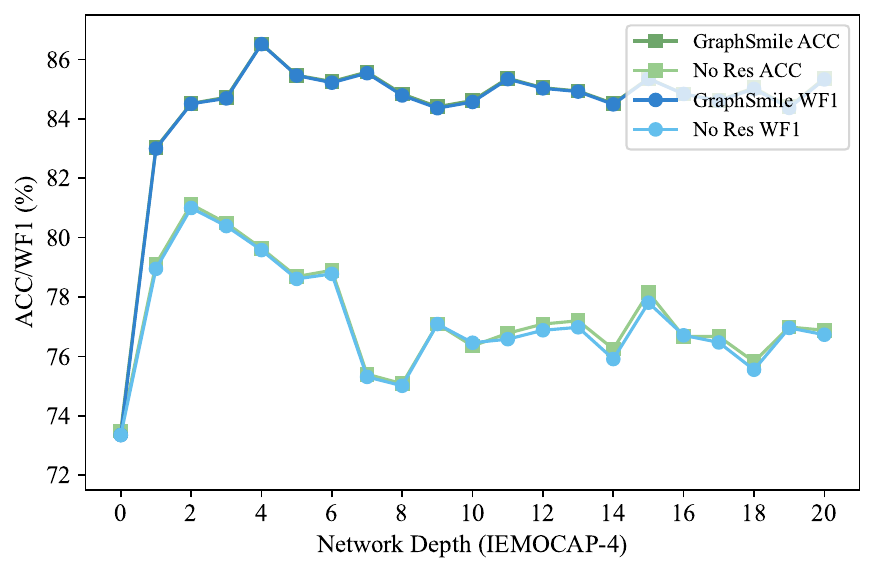}%
\label{fig:network_depth_iemocap4}}
\vfil
{\includegraphics[width=0.45\linewidth]{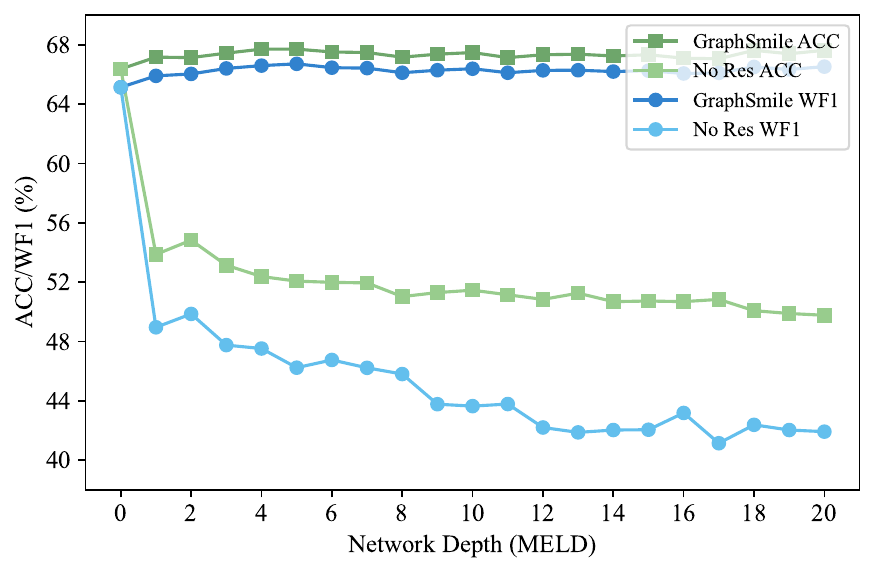}%
\label{fig:network_depth_meld}}
\hfil
{\includegraphics[width=0.45\linewidth]{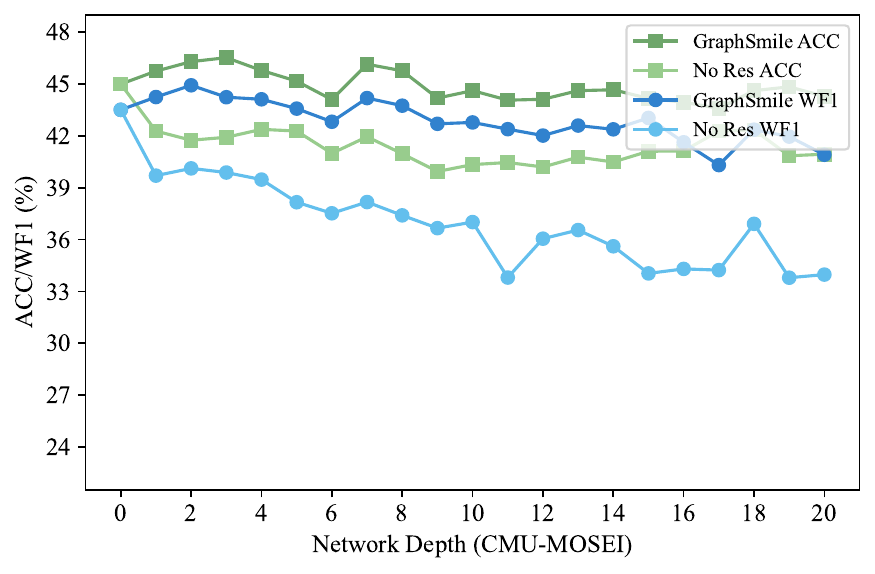}%
\label{fig:network_depth_cmumosei}}
\caption{Performance variations of GraphSmile for the MERC task with the network depth. The abbreviation ``Res'' indicates the residual connection operation.}
\label{fig:network_depth}
\end{figure*}

Early GNNs commonly demonstrate the notorious over-smoothing phenomenon~\cite{10.1109/TPAMI.2021.3074057}. This issue refers to the gradual homogenization of node features as network depth increases, leading to a sharp decline in model performance after reaching a certain threshold. In Fig.~\ref{fig:network_depth}, the dark blue and dark green lines intuitively display the performance variation of GraphSmile for the MERC task with network depth. On the IEMOCAP (6-way) dataset, GraphSmile's performance improves with the increase of network layers, slightly declines after reaching the peak, and then maintains a slight fluctuation. The performance changes on the IEMOCAP (4-way) and MELD datasets exhibit a similar pattern. These phenomena indicate that GraphSmile successfully circumvents the over-smoothing issue. In Fig.~\ref{fig:network_depth}, we also illustrate the performance variation of the model without residual connections (i.e., not employing Eq.~\ref{eq:res_connect}) adopting light blue and light green lines. It can be observed that as the network depth increases, the performance of this model (i.e., the incomplete GraphSmile) rapidly declines after reaching a certain threshold, showing signs of over-smoothing. The performance comparison between the complete and incomplete versions substantiates the effectiveness of residual connections in alleviating the over-smoothing problem. Despite GraphSmile is less effective against over-smoothing on the CMU-MOSEI dataset due to the higher classification difficulty, the complete model still indicates a significant performance advantage over the incomplete version.

\subsection{Impact of Window Size}
\begin{figure*}[htbp]
\centering
{\includegraphics[width=0.45\linewidth]{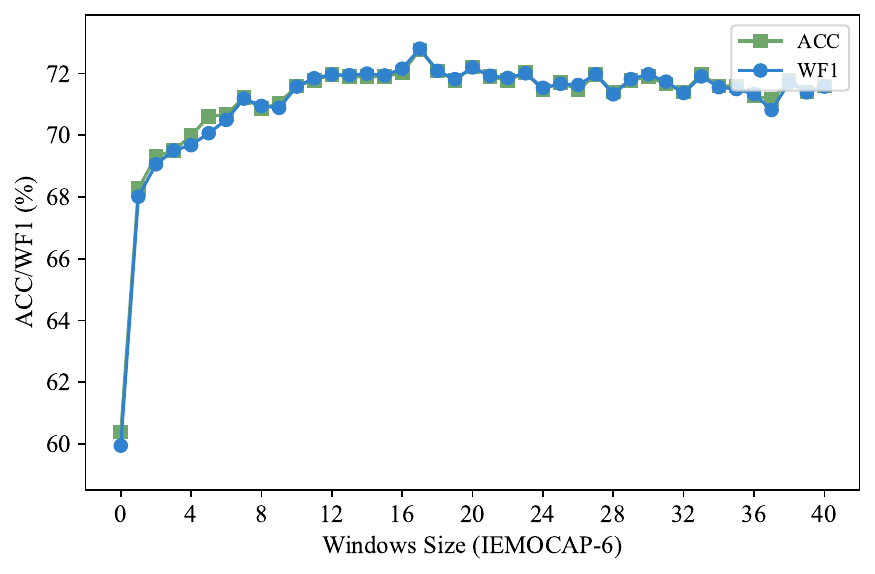}%
\label{fig:window_size_iemocap6}}
\hfil
{\includegraphics[width=0.45\linewidth]{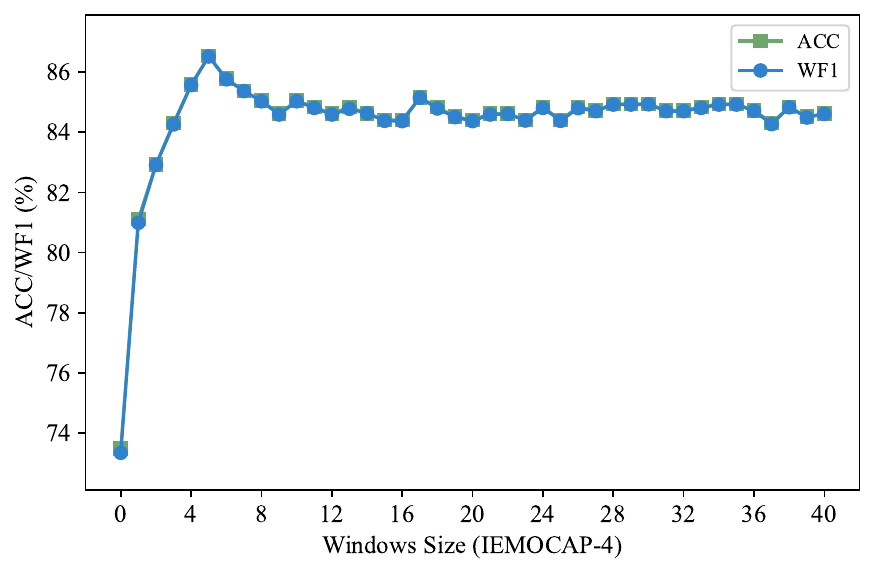}%
\label{fig:window_size_iemocap4}}
\vfil
{\includegraphics[width=0.45\linewidth]{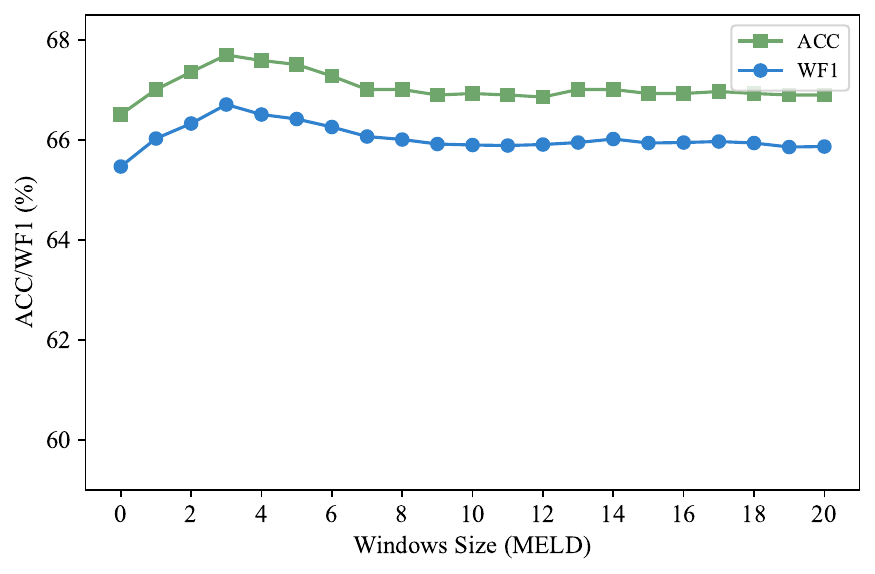}%
\label{fig:window_size_meld}}
\hfil
{\includegraphics[width=0.45\linewidth]{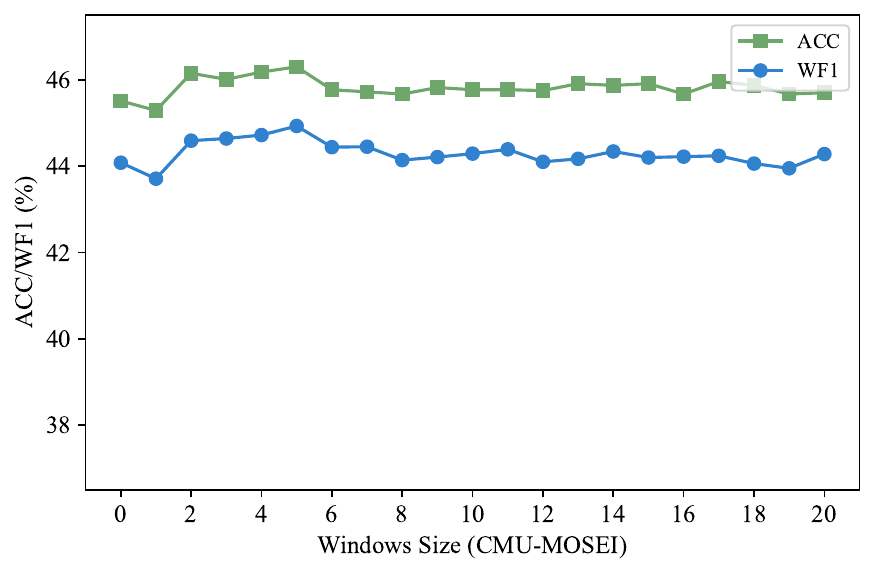}%
\label{fig:window_size_cmumosei}}
\caption{Performance variations of GraphSmile for the MERC task with the window size. Considering the average dialogue lengths, we set different size ranges for these datasets.}
\label{fig:window_size}
\end{figure*}

During the construction of multimodal dialogue graphs, we introduce two sliding windows to eliminate irrelevant relationships. For ease of experimentation, we employ two equal-length windows, i.e., setting the sizes of the past and future windows to be equal. Fig.~\ref{fig:window_size} details the performance of GraphSmile for the MERC task under different window sizes. Taking the IEMOCAP (6-way) dataset as an example, when the window size is 0 (i.e., without considering contextual information), the performance of GraphSmile drops to the lowest, which is also reflected in the IEMOCAP (4-way) and MELD datasets. These results indicate that context modeling plays a crucial role in the ERC task. As expected, the performance trends across all datasets exhibit a pattern of initial increase followed by a decrease as the window size increases, eventually fluctuating slightly within a certain range. This suggests that there is an optimal window size that allows the model to capture sufficient contextual information while avoiding the introduction of excessive noise.

\subsection{Embedding Visualization}
\begin{figure*}[htbp]
\centering
\subfloat[Initial (IEMOCAP-6)]{\includegraphics[width=0.24\linewidth]{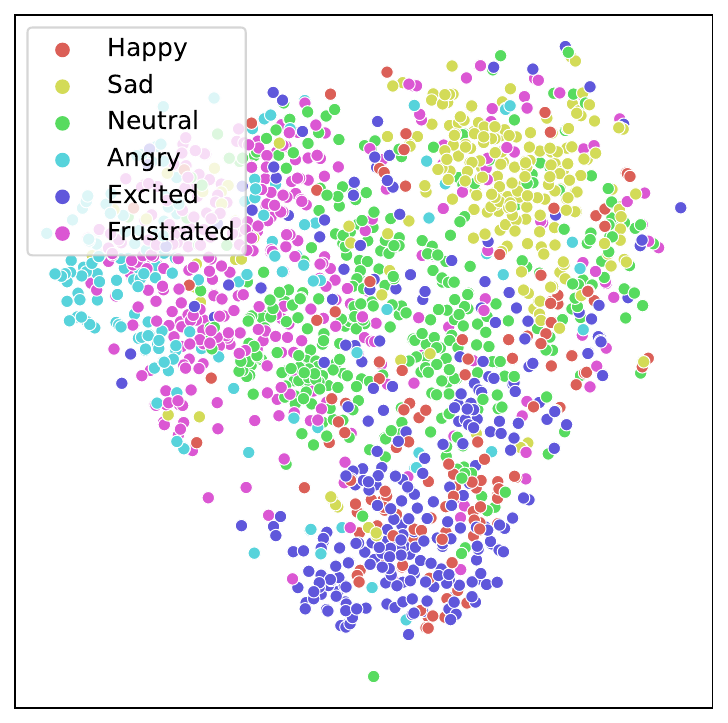}%
\label{fig:embed_visual_emo_initial_iemocap6}}
\hfil
\subfloat[MMGCN (IEMOCAP-6)]{\includegraphics[width=0.24\linewidth]{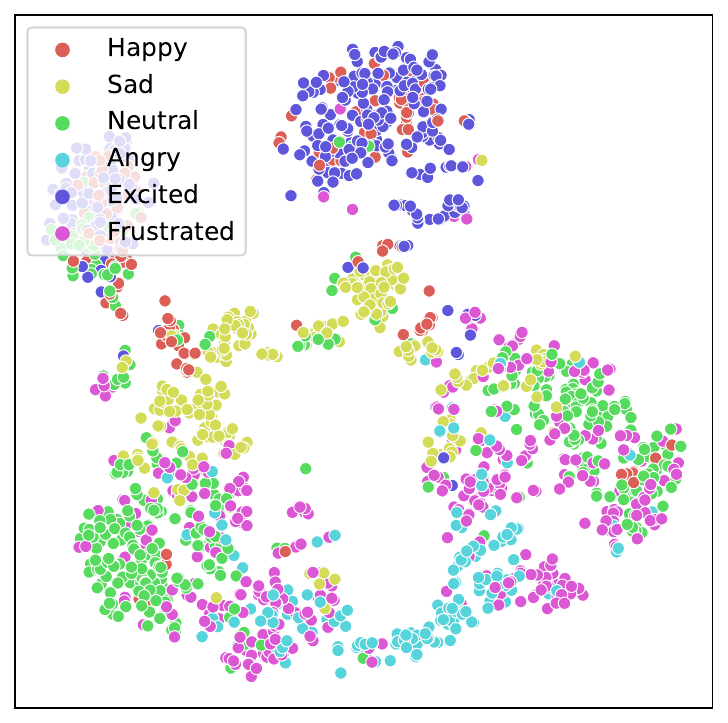}%
\label{fig:embed_visual_emo_mmgcn_iemocap6}}
\hfil
\subfloat[M3Net (IEMOCAP-6)]{\includegraphics[width=0.24\linewidth]{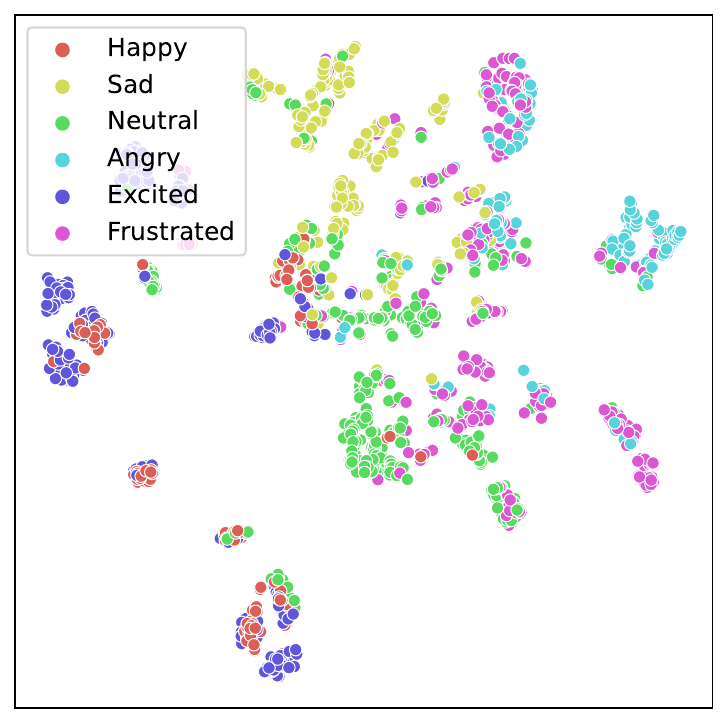}%
\label{fig:embed_visual_emo_m3net_iemocap6}}
\hfil
\subfloat[GraphSmile (IEMOCAP-6)]{\includegraphics[width=0.24\linewidth]{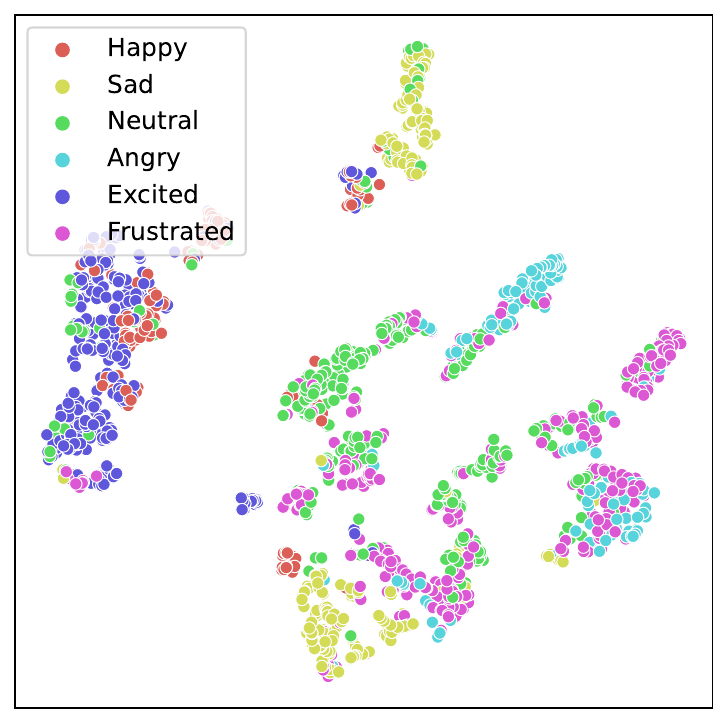}%
\label{fig:embed_visual_emo_graphsmile_iemocap6}}
\vfil
\subfloat[Initial (IEMOCAP-4)]{\includegraphics[width=0.24\linewidth]{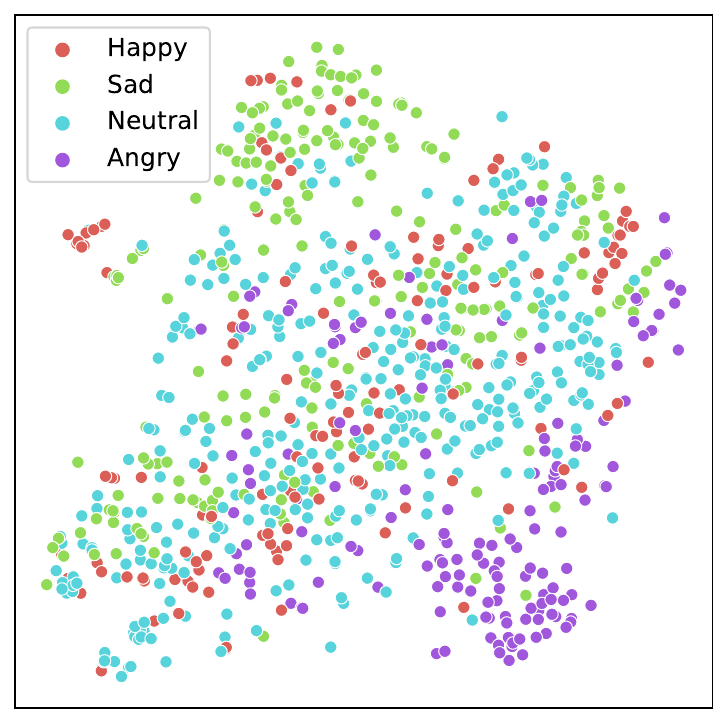}%
\label{fig:embed_visual_emo_initial_iemocap4}}
\hfil
\subfloat[MMGCN (IEMOCAP-4)]{\includegraphics[width=0.24\linewidth]{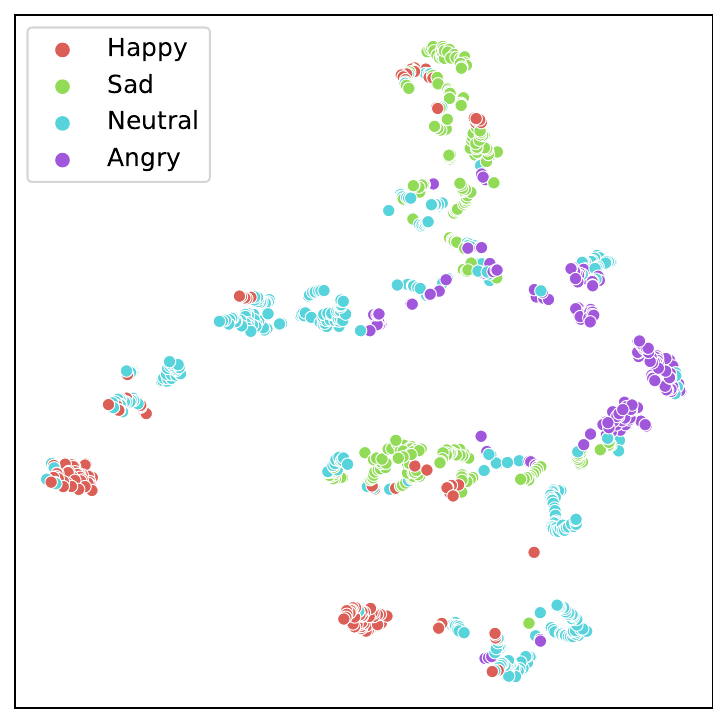}%
\label{fig:embed_visual_emo_mmgcn_iemocap4}}
\hfil
\subfloat[M3Net (IEMOCAP-4)]{\includegraphics[width=0.24\linewidth]{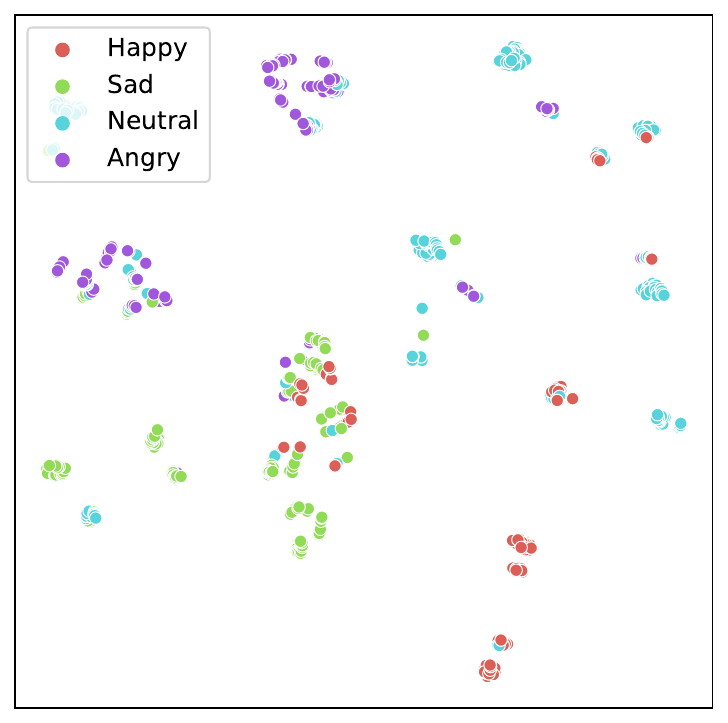}%
\label{fig:embed_visual_emo_m3net_iemocap4}}
\hfil
\subfloat[GraphSmile (IEMOCAP-4)]{\includegraphics[width=0.24\linewidth]{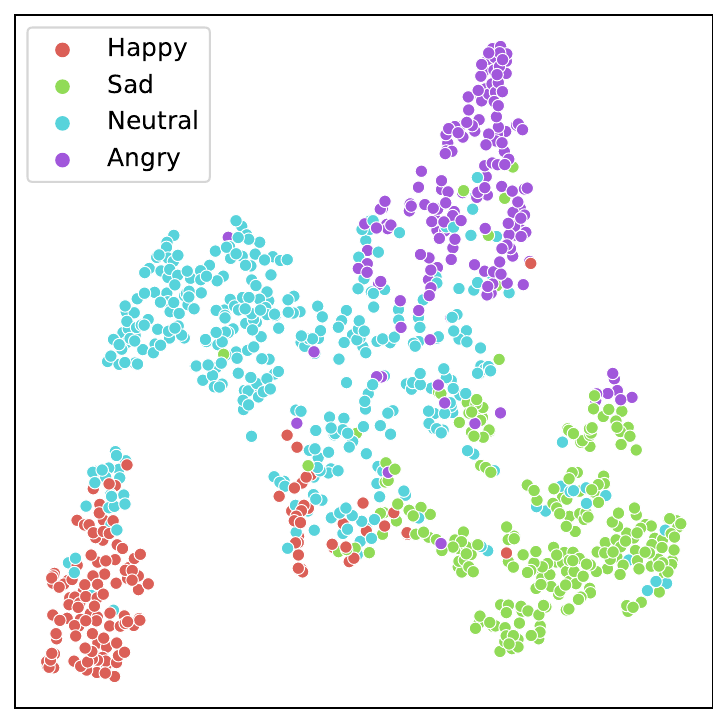}%
\label{fig:embed_visual_emo_graphsmile_iemocap4}}
\vfil
\subfloat[Initial (MELD)]{\includegraphics[width=0.24\linewidth]{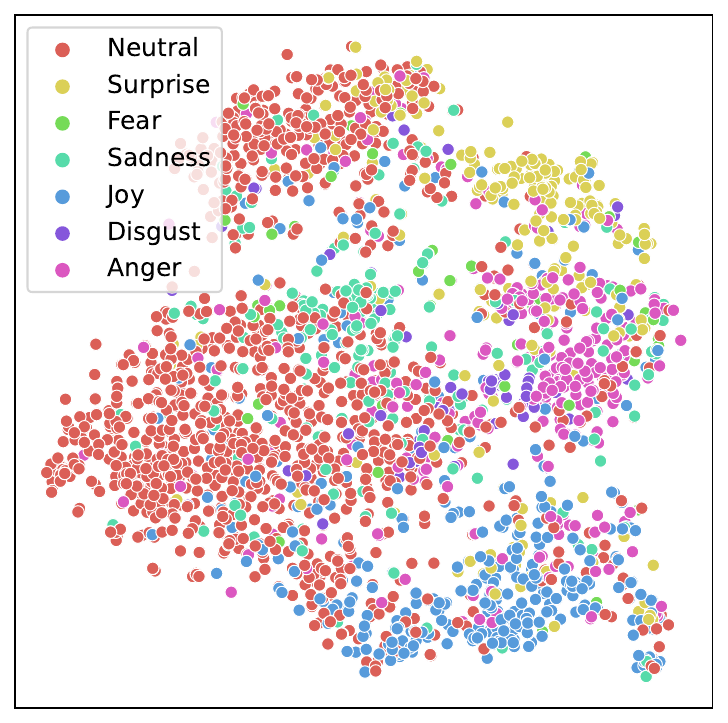}%
\label{fig:embed_visual_emo_initial_meld}}
\hfil
\subfloat[MMGCN (MELD)]{\includegraphics[width=0.24\linewidth]{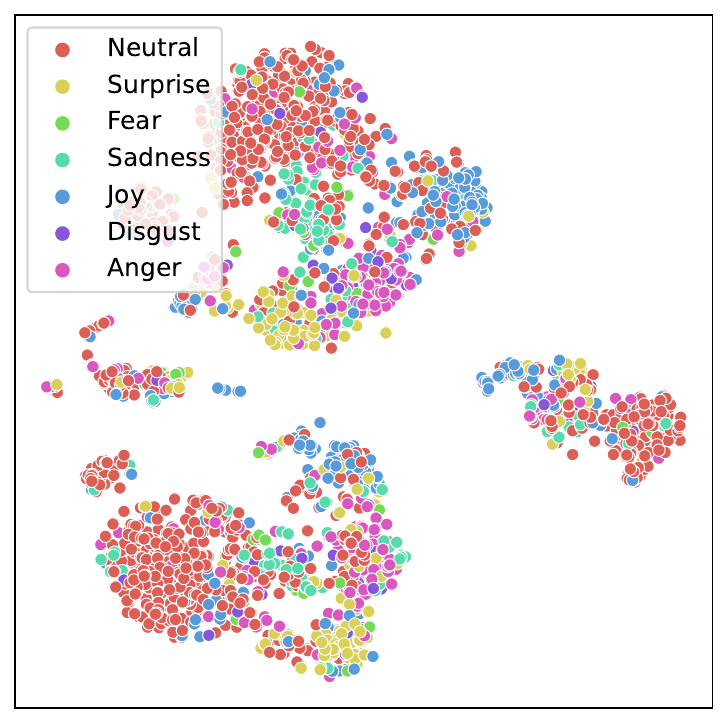}%
\label{fig:embed_visual_emo_mmgcn_meld}}
\hfil
\subfloat[M3Net (MELD)]{\includegraphics[width=0.24\linewidth]{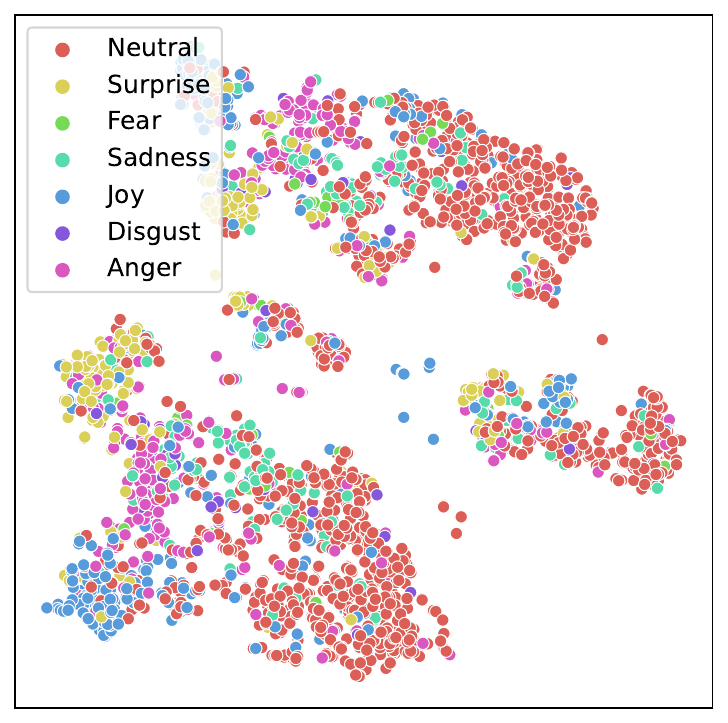}%
\label{fig:embed_visual_emo_m3net_meld}}
\hfil
\subfloat[GraphSmile (MELD)]{\includegraphics[width=0.24\linewidth]{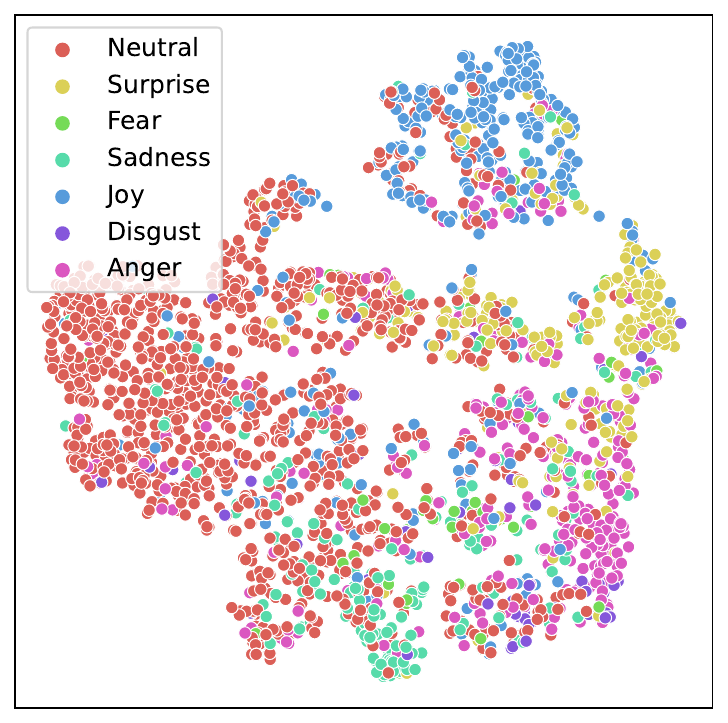}%
\label{fig:embed_visual_emo_graphsmile_meld}}
\vfil
\subfloat[Initial (CMU-MOSEI)]{\includegraphics[width=0.24\linewidth]{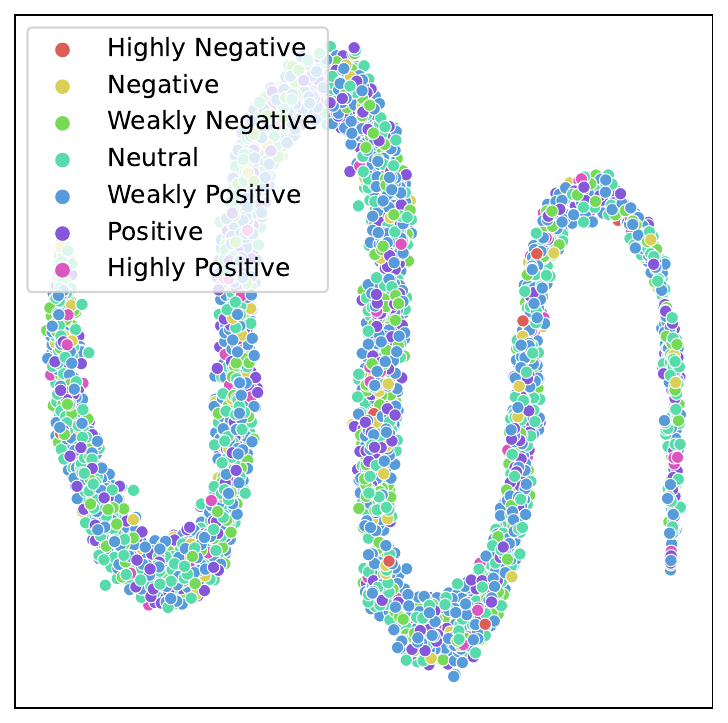}%
\label{fig:embed_visual_emo_initial_cmumosei7}}
\hfil
\subfloat[MMGCN (CMU-MOSEI)]{\includegraphics[width=0.24\linewidth]{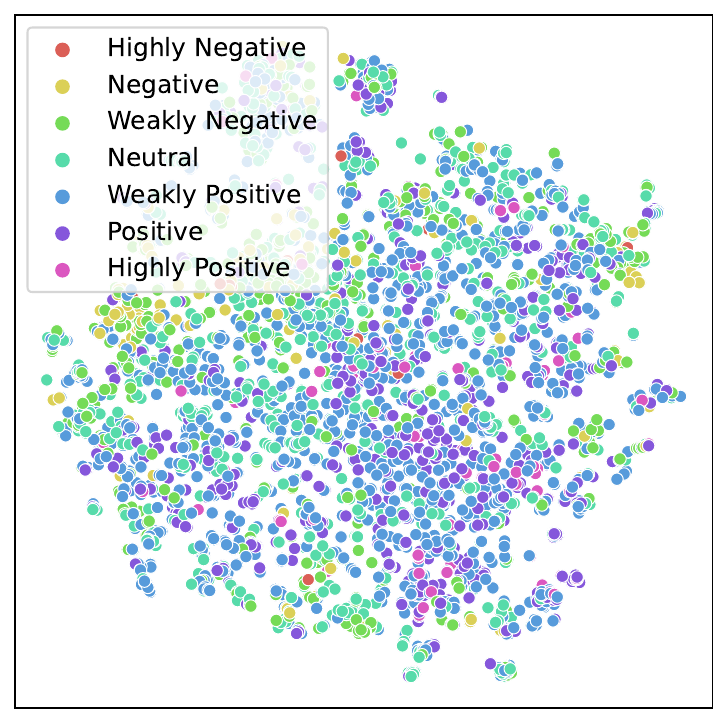}%
\label{fig:embed_visual_emo_mmgcn_cmumosei7}}
\hfil
\subfloat[M3Net (CMU-MOSEI)]{\includegraphics[width=0.24\linewidth]{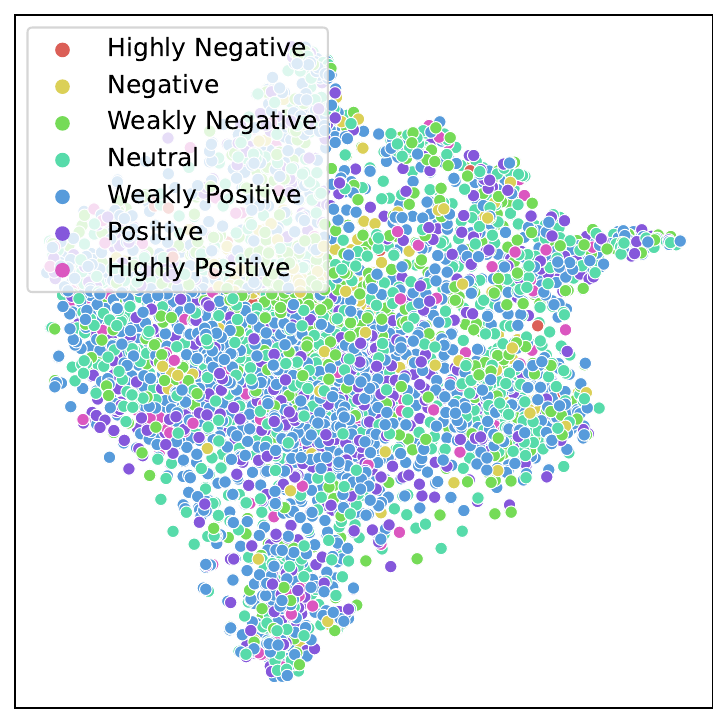}%
\label{fig:embed_visual_emo_m3net_cmumosei7}}
\hfil
\subfloat[GraphSmile (CMU-MOSEI)]{\includegraphics[width=0.24\linewidth]{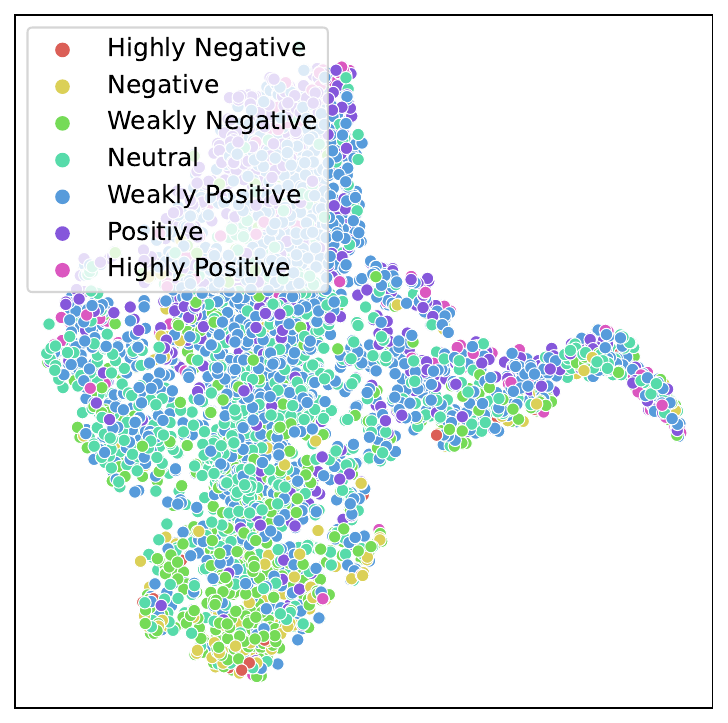}%
\label{fig:embed_visual_emo_graphsmile_cmumosei7}}
\caption{Visualization of utterance embeddings for all datasets. Here, ``Initial (IEMOCAP-6)" signifies the t-SNE visualization of the initial features in the IEMOCAP (6-way) dataset, and ``MMGCN (IEMOCAP-6)" indicates the visualization of the embeddings extracted by MMGCN in the same dataset. Similar nomenclature is applied to other figures.}
\label{fig:embed_visual_emo}
\end{figure*}
To visually display the classification efficacy of GraphSmile for the MERC task, we employ T-SNE~\cite{JMLR:v9:vandermaaten08a} to visualize the utterance embeddings. As depicted in Fig.~\ref{fig:embed_visual_emo}, GraphSmile effectively segregates samples of distinct emotions in the IEMOCAP dataset, thereby substantiating its potency in emotion classification. Although MMGCN can to some extent differentiate emotional samples, its performance does not match that of our proposed model. Similar to GraphSmile, M3Net achieves commendable classification outcomes on this dataset. Analogous phenomena are observed in the MELD dataset. Furthermore, in contrast to the performance on the IEMOCAP and MELD datasets, GraphSmile exhibits suboptimal results on the CMU-MOSEI dataset, aligning with the outcomes of MMGCN and M3Net.

\subsection{Limitation}
\begin{figure*}[htbp]
\centering
{\includegraphics[width=0.45\linewidth]{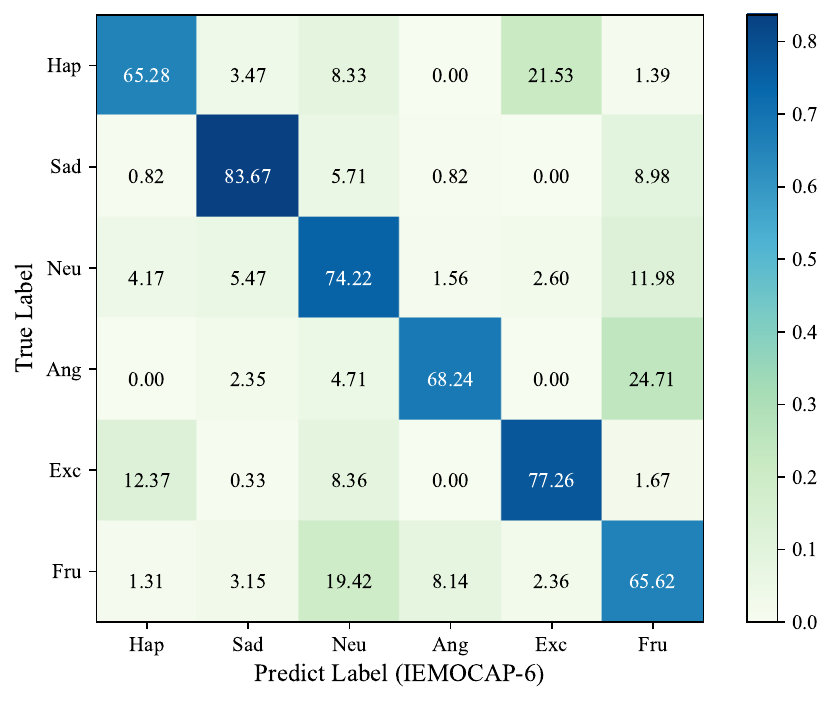}%
\label{fig:confusion_matrix_iemocap6}}
\hfil
{\includegraphics[width=0.45\linewidth]{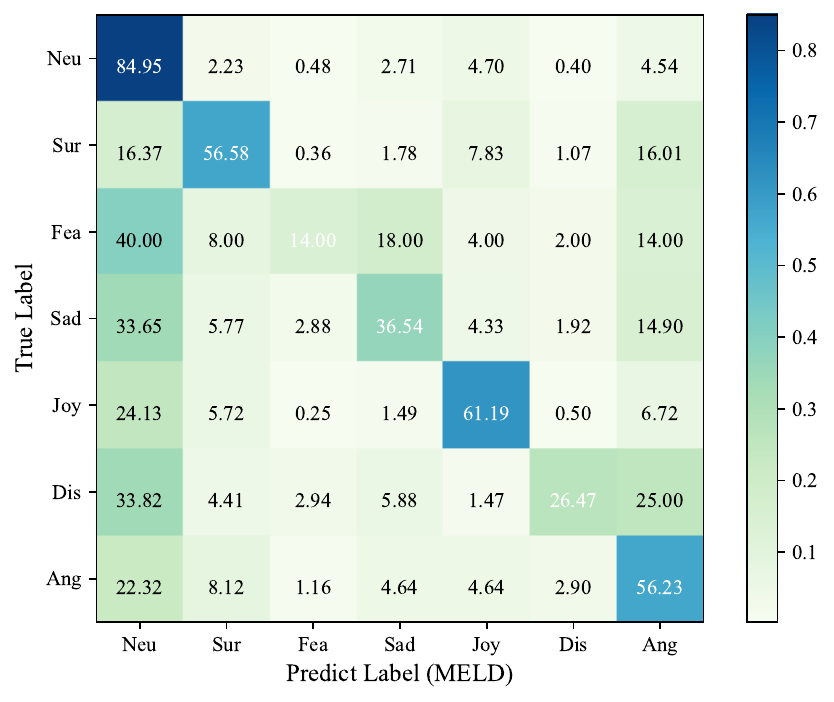}%
\label{fig:confusion_matrix_meld}}
\caption{Confusion matrices on the IEMOCAP (6-way) and MELD datasets. The abbreviation ``Hap'' denotes Happy and other emotions by analogy.}
\label{fig:confusion_matrix}
\end{figure*}
Like much of the previous work, GraphSmile has the limitation when dealing with similar emotions. As shown in Fig.~\ref{fig:confusion_matrix}, on the IEMOCAP (6-way) dataset, Happy samples have a high probability of being identified as the positive emotion Excited. Conversely, Excited samples also have a certain probability of being classified as Happy. Another limitation of GraphSmile is its difficulty in correctly detecting minority class samples on the extremely imbalanced dataset. For instance, on the MELD dataset, samples of Fear and Disgust, which are extreme minority classes, are challenging to classify correctly and tend to be identified as the majority class Neutral. Additionally, visual information (e.g., facial expressions) and acoustic information (e.g., intonation) should theoretically contain significant emotional cues. However, as Table~\ref{tab:ablation_study} reflects, due to the presence of a large amount of noise, they do not leverage their unique advantages in dialogue affective tasks. This may lead GraphSmile to favor the textual modality in cross-modal modeling, thereby reducing the effective mining of visual and acoustic information.

\section{Conclusion}\label{sec:conclusion}
In this research, we propose an innovative multitask multimodal affective model named GraphSmile, specifically tailored for MERC and MSAC tasks. The proposed model achieves synergistic optimization of these tasks by integrating emotion recognition, sentiment analysis, and sentiment shift detection, thereby delving into the intricate emotional and sentimental cues embedded in multimodal dialogues. GraphSmile employs a unique graph construction manner that not only facilitates comprehensive capture of cross-modal interaction cues but also effectively mitigates redundant connections in the multimodal dialogue graph. To circumvent potential conflicts arising from the fusion of multi-source data, we design a graph structure based multimodal fusion module, termed GSF. This module adeptly harnesses the graph structure, optimizing the integration process by alternately absorbing inter-modal and intra-modal affective dependencies in a layered fashion. In the GSF module, we introduce a residual connection mechanism integrated with fully connected layers, which not only alleviates the over-smoothing issue common in GNNs but also significantly enhances the model's capacity for affective expression. Furthermore, to bolster the model's capability to capture the sentiment dynamics, we develop a plug-and-play auxiliary module known as SDP. This module explicitly models the sentimental transitions between utterances, augmenting the model's discernment of sentimental variances. We have conducted extensive validation experiments on several public benchmark datasets, and the results robustly substantiate the efficacy of GraphSmile and its components.

In future work, we intend to address other salient challenges in the realm of affective computing in conversation. For instance, (1) tackling the differentiation of similar emotions and issues of class imbalance through contrastive learning strategies, and (2) delving deeper into the sentimental dynamics in conversation by considering the temporal logic of transitions and refining the types of shift labels. We will also explore more sophisticated multimodal fusion strategies to capitalize on the unique strengths of each modality and enhance the model's robustness against noise and irrelevant information. Recently, large language models (LLMs) have demonstrated formidable reasoning capabilities, thus integrating MERC and MSAC with LLM technologies will further propel the development of affective community.

\balance
\bibliographystyle{IEEEtran}
\bibliography{IEEEabrv,graphsmile.bib}
% \balance
% \newpage

\end{document}